\documentclass{article}

\usepackage[main, final]{neurips_2026}

\usepackage[utf8]{inputenc} %
\usepackage[T1]{fontenc}    %
\usepackage{hyperref}       %
\usepackage{url}            %
\usepackage{booktabs}       %
\usepackage{amsfonts}       %
\usepackage{nicefrac}       %
\usepackage{microtype}      %
\usepackage{xcolor}         %
\usepackage{graphicx}
\usepackage{amsmath}
\usepackage{amssymb}
\usepackage[textsize=tiny]{todonotes}
\usepackage{booktabs, multirow}
\usepackage{wrapfig}
\usepackage{float}
\usepackage{needspace}
\usepackage{placeins}
\usepackage{caption}
\usepackage{titletoc}
\usepackage{subcaption}

\usepackage{xcolor}
\usepackage{tcolorbox}
\tcbuselibrary{skins,breakable}

\usepackage{todonotes}
\usepackage{soul}

\definecolor{myolive}{RGB}{110,120,35}
\definecolor{mypurple}{RGB}{135,65,160}
\definecolor{myblue}{RGB}{35,100,185}

\soulregister\citep7
\soulregister\citet7
\soulregister\ref7
\soulregister\eqref7
\soulregister\emph7
\soulregister\textbf7
\soulregister\textit7

\NewDocumentCommand{\authorcomment}{m m o m}{%
  \IfNoValueTF{#3}{%
    \todo[
      line,
      color=#1!15,
      linecolor=#1,
      bordercolor=#1
    ]{#2: #4}%
  }{%
    \begingroup
      \color{#1}%
      [%
      \color{black}%
      \ul{#3}%
      \color{#1}%
      ]%
    \endgroup
    \todo[
      line,
      color=#1!15,
      linecolor=#1,
      bordercolor=#1
    ]{#2: #4}%
  }%
}

\NewDocumentCommand{\isra}{o m}{%
  \IfNoValueTF{#1}
    {\authorcomment{red!70!black}{IS}{#2}}
    {\authorcomment{red!70!black}{IS}[#1]{#2}}%
}

\NewDocumentCommand{\constanza}{o m}{%
  \IfNoValueTF{#1}
    {\authorcomment{myolive}{CF}{#2}}
    {\authorcomment{myolive}{CF}[#1]{#2}}%
}

\NewDocumentCommand{\stella}{o m}{%
  \IfNoValueTF{#1}
    {\authorcomment{green!60!black}{SF}{#2}}
    {\authorcomment{green!60!black}{SF}[#1]{#2}}%
}

\NewDocumentCommand{\dan}{o m}{%
  \IfNoValueTF{#1}
    {\authorcomment{mypurple}{DO}{#2}}
    {\authorcomment{mypurple}{DO}[#1]{#2}}%
}

\NewDocumentCommand{\des}{o m}{%
  \IfNoValueTF{#1}
    {\authorcomment{myblue}{DE}{#2}}
    {\authorcomment{myblue}{DE}[#1]{#2}}%
}

\definecolor{last_token_color}{HTML}{96c458}
\definecolor{image_patch_color}{HTML}{85b7eb}
\definecolor{text_patch_color}{HTML}{ef9f28}
\renewcommand{\hl}[2]{{\setlength{\fboxsep}{1pt}\colorbox{#1}{\smash[b]{#2}}}}

\title{Pathways of Visual Information Flow in Vision-Language Models}

\author{
  Israfel Salazar$^{1}$ \quad
  Stella Frank$^{2}$ \quad
  Dan Oneata$^{3, 4}$ \quad
  Desmond Elliott$^{1}$ \quad
  Constanza Fierro$^{1}$ \\
  $^{1}$University of Copenhagen \quad \\
  $^{2}$Technical University of Denmark \quad \\
  $^{3}$POLITEHNICA Bucharest \quad $^{4}$Bitdefender, Romania
  }

\begin{document}

\maketitle

\begin{abstract}

We study how visual information is routed in vision-language models (VLMs). Using causal patching on controlled synthetic and natural datasets, we find that models rely on two distinct pathways to solve visual tasks: A \emph{direct pathway}, where visual information is retained in image token representations and read out by the final token at later layers, and a \emph{text-mediated pathway}, where visual information is first transferred to the query tokens and then read out by the final token. Across three visual tasks, we show that pathway selection is task-dependent, and that data distribution and prompt design can also modulate which pathway is used to solve the image-based query. Moreover, using attention knockouts and corrupted-input patching, we find that these pathways are flexible, under certain interventions, models can rely on the text-mediated pathway as a fallback when the usual pathway is ablated. This behavior unifies findings in prior work and shows that ablation-based interventions can reveal what models could do rather than what they normally do. Together, our results provide a mechanistic characterization of visual information flow in VLMs and highlight the flexibility of their internal mechanisms under intervention.\footnote{
Code: {\scriptsize\url{https://github.com/israfelsr/vlm-pathways}}. Correspondence: Israfel Salazar <\href{mailto:israfel.salazar@di.ku.dk}{israfel.salazar@di.ku.dk}>
}

\end{abstract}

\section{Introduction}
\label{sec:intro}

Vision-language models can solve complex multimodal tasks that involve understanding image and text input, yet it remains unclear how visual information is combined in the language model during text generation.
Understanding this information flow is critical for interpretability \citep{basu2024understanding}, trustworthiness \citep{sharkey2025open}, and test-time interventions~\citep{chen2025spatial}.
Recent work has made progress in understanding how visual information is integrated in Generative Visual-Language Models (VLMs), mostly relying on attention score analysis \citep{chen2025spatial,kaduri2025s} or attention interventions \citep{neo2024towards,zhang2025cross}, where a set of tokens is prevented from attending to another set of tokens (e.g.~blocking the last token from attending to the image).

These works point to different mechanisms of visual information flow. \citet{neo2024towards} observed that for image recognition, the visual information flows directly from the image tokens to the last token, whereas \citet{zhang2025cross} found that across different tasks, the text query always mediates the flow of visual information between the image tokens and the last token. Both studies, however, are based on attention ablations, which can only tell us whether the model can produce the correct answer without a given component, but not necessarily whether it typically \textit{uses} that component to generate the answer. More recently, a causal intervention study supported text-mediated visual information flow in VLMs, although the analysis was limited to a single task~\citep{kang2026linear}. As a result, there is still limited understanding of how image-based information flows in VLMs when solving image-based queries. 

\begin{figure}[t]
  \centering
  \includegraphics[width=0.8\textwidth]{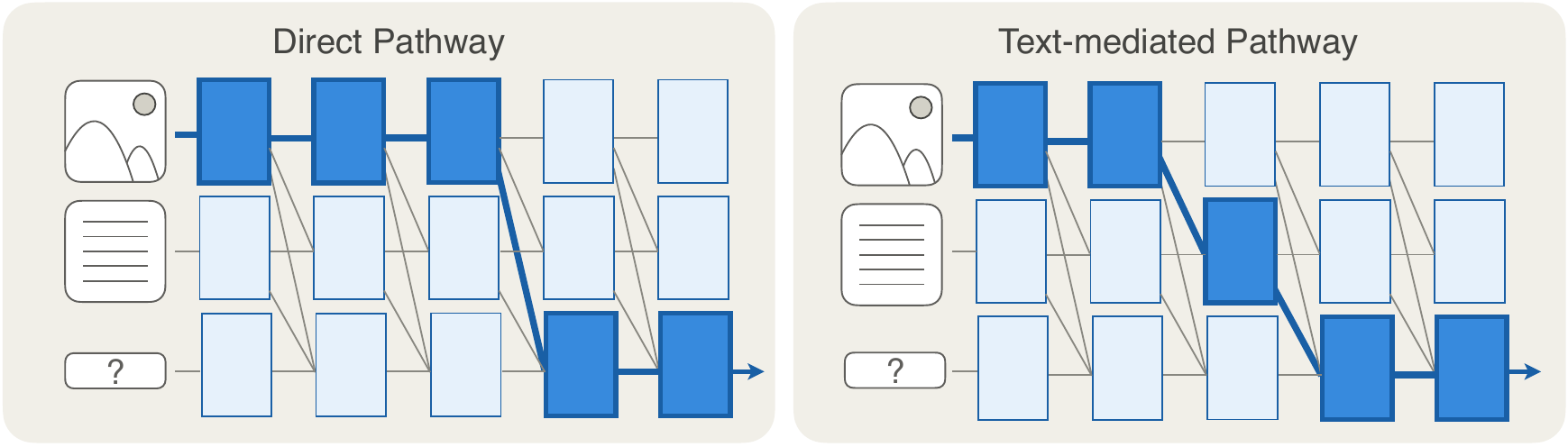}
  \caption{Illustration of the two pathways VLMs rely on for routing visual information. In the direct pathway, the final token attends to image tokens to extract visual information. In the text-mediated pathway, visual information is first transferred to the query tokens, which are then read out by the final token. These pathways provide alternative routes for integrating visual information.}
  \label{fig:pathways_main}
\end{figure}

To investigate how visual information reaches the last token, we rely on causal patching ~\citep{geiger2025causal} with counterfactual examples across both synthetic and natural datasets. Our experiments cover three visual tasks requiring different levels of visual grounding and understanding: object recognition (identifying entities), predicting spatial relations (reasoning over spatial positions between entities), and object localization (mapping entities to positions). We find across model families (Qwen3~\citep{bai2025qwen3}, InternVL3.5~\citep{wang2025internvl3}, and  LLaVA-1.5~\citep{liu2024improved}) that VLMs solve these tasks by routing the visual information in two distinct pathways (Figure~\ref{fig:pathways_main}): a direct pathway, where the last token directly reads the visual information from the image tokens, and a text-mediated pathway, where visual information is transferred to the query tokens representations, which are then read out by the last token.

Unifying previous works, we find that pathway selection is task-dependent: VLMs solve spatial relationships through text-mediation, but use the direct pathway for object recognition.
Within a given setting the model consistently uses a single pathway, but the choice of pathway can depend on data distribution or prompt design.
For instance, for localization queries, models use the direct pathway on synthetic data but switch to text-mediation on natural datasets.

Beyond characterizing these pathways, we show they coexist as backup alternatives. When we disrupt the direct pathway through attention knockouts or corrupted patching, models reroute through text-mediation and maintain task performance. In text-only language models, ``backup'' circuits partially compensate when primary components are ablated, but with degraded performance \citep{wang2023interpretability, mcgrath2023hydra, mcdougall-etal-2024-copy}. We observe a stronger form of this phenomenon in VLMs, with models switching routing strategies and maintaining task performance through a different pathway. This has direct methodological implications, as interventions meant to reveal how VLMs process visual information can instead trigger these fallback pathways, producing conclusions that reflect the intervention regime rather than usual model behavior.

Our main contributions are:
\begin{itemize}
    \item We identify and characterize two different pathways that VLMs use to channel visual information from the image tokens to the last token (Figure~\ref{fig:pathways_main}) using both synthetic and natural data.
    \item We show that pathway selection is task-dependent and modulated by data distribution and prompt design, with each task and data type consistently using a single routing strategy.
    \item We find that these pathways are not rigid: models can flexibly reroute through text-mediation when the direct pathway is disrupted. 
    This implies that standard interventions on VLMs risk activating fallback pathways, which may lead to conclusions that conflate the mechanisms a model uses with those it could use.
\end{itemize}

\begin{figure}[t]
  \centering
  \includegraphics[width=1\textwidth]{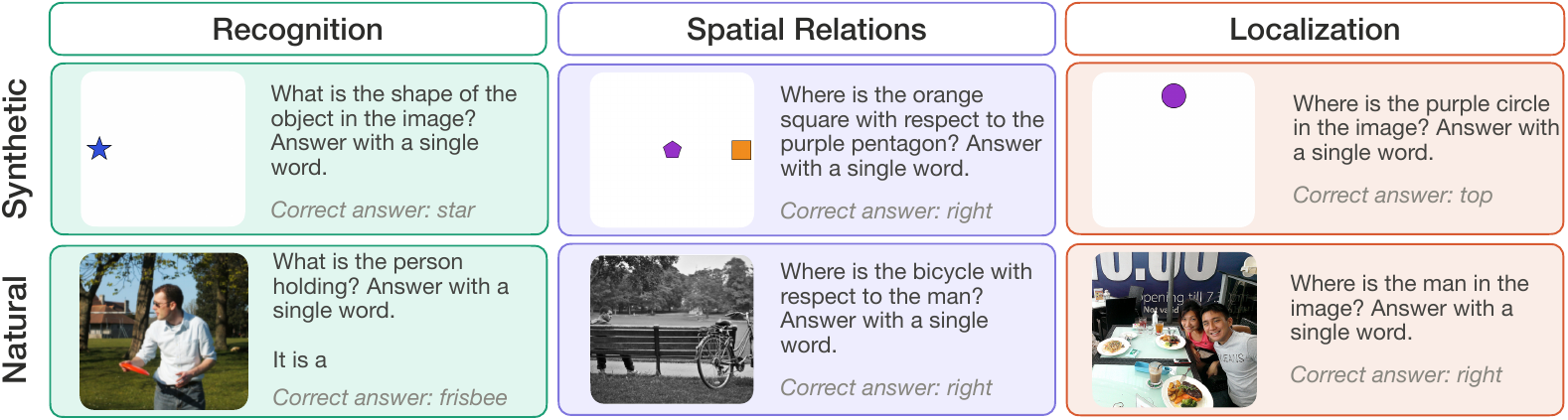}
  \caption{Task definitions and examples. We consider three visual tasks: object recognition, spatial relations, and localization. We evaluate them in both synthetic (top) and natural images (bottom).}
  \label{fig:tasks_samples}
\end{figure}

\section{Methodology}
\label{sec:experimental_setup}
To characterize how visual information is routed through VLMs, our methodology is built around four design components: (i) a set of tasks covering different types of visual understanding, (ii) data that enables both controlled experiments and test generalization, (iii) interventions that distinguish what a model could use from what it actually uses, and (iv) evaluation across multiple model families. 

\paragraph{Tasks.}
We select three different types of visual understanding that cover core perceptual abilities: (1) recognition, in which models need to identify an object in the image; (2) spatial relations, where the model identifies the position of one object with respect to another; and (3) localization, where the model extracts the global position of a specific object from the viewer's point of view. We study these tasks under a unified methodology to test whether different understanding types reflect a single shared mechanism or if they induce distinct types of information flow.

\paragraph{Data: Synthetic Shapes.} We create one synthetic dataset per task using coloured shapes (Figure~\ref{fig:samples_syn_stacked}). For recognition and localization a single shape is placed at the top, bottom, left, or right, isolating object identification and absolute position respectively. For relations, the same layout is used but a second shape is added at the center. Because all three datasets share the same primitives and spatial grid, we can study visual information flow in a controlled manner. Each dataset contains 400 examples, including 6 different shapes and 6 colors. More details and samples in Appendix~\ref{sec:app_datasets_synth}.

\paragraph{Data: Natural.} We also evaluate on natural images to determine whether the mechanisms identified in the synthetic data generalize beyond controlled environments. For localization and spatial relations we include images from five different datasets: What's Up~\citep{kamath2023s}, which consists of structured two-object scenes, two natural splits from COCO~\citep{lin2014microsoft}, Visual Genome~\citep[VG]{krishna2017visual}, and VSR~\citep{liu2023visual} (see examples in Figure~\ref{fig:samples_nat_stacked} and Appendix~\ref{sec:app_datasets_nat} for more details). For recognition, we use COCO images with the queries from~\citet{neo2024towards}. Across all the natural datasets, the task definition is the same as in the synthetic setup while increasing visual complexity. We report results aggregated across datasets for each of the three tasks, with per-dataset results and analysis provided in Appendix~\ref{sec:app_per_dataset}. 

\paragraph{Generation and Output Constraints.}
We evaluate under two prompting protocols, \textit{choices} and \textit{open}, which differ in how they present and constrain the answer space. In the \textit{choices} setting, the model is prompted to reply with a specific word from a fixed set of candidate answers (e.g. ``Answer only with above, below, left, right'', see Appendix~\ref{sec:app_prompts} for all the prompts). In this setting, accuracy is computed using the top1 predicted token. In the \textit{open} setting, no candidate set is provided and the model is instructed to produce a single-word response (Figure~\ref{fig:tasks_samples}). We evaluate by restricting the argmax to the set of valid answer tokens (the prepositions for spatial tasks and the object labels for recognition). 
The \textit{open} setting is our primary evaluation mode, and the constrained \textit{choices} generation is reported alongside to study the effect of multiple choices on model behavior. 
For natural recognition, we further standardize the output format by prefixing the prompt with ``It is a'', to enforce direct object naming and reduce variability. We additionally evaluate prefilled multi-token answers as a proxy for multi-token generation (see Appendix \ref{sec:app_multitoken}).

\paragraph{Notation: Information Pathways.}
To study how information flows from the image tokens to the final token we consider a set of candidate pathways. Let \((x_1, ..., x_n)\) be the input tokens. Excluding special tokens, let $\mathcal{I}$ and $\mathcal{T}$ denote the sets of image and text tokens, respectively, excluding the final token $x_n$ , which is treated as a singleton. We represent information flow using pairs of tokens \((x_i,x_j)\) , where token $x_j$ attends to token $x_i$ . We consider three pathways (or routes) $r$:
\begin{equation}
\begin{gathered}
    r_{\mathcal{I} \to \text{last}} = \{(x_i, x_n) : x_i \in \mathcal{I}\}, \\ 
    r_{\mathcal{I} \to \mathcal{T}} = \{(x_i, x_t) : x_i \in \mathcal{I}, x_t \in \mathcal{T}\}, \quad
    r_{\mathcal{T} \to \text{last}} = \{(x_t, x_n) : x_t \in \mathcal{T}\}
\end{gathered}
\end{equation}
These correspond to two candidate routing mechanisms for visual information: (i) a direct pathway ($r_{\mathcal{I} \to \text{last}}$) where the last token directly attends to image tokens to extract visual information, and a text-mediated pathway, where visual information is first transferred from image tokens to query tokens ($r_{\mathcal{I} \to \mathcal{T}}$), and then read out by the last token through $r_{\mathcal{T} \to \text{last}}$.  Sets $\mathcal{I}$ and $\mathcal{T}$ exclude special tokens (Appendix~\ref{sec:app_prompts}). To test when these pathways are used we perform the following interventions.

\paragraph{Intervention: Attention Knockouts.}
We analyze the importance of information flow between two tokens by suppressing the attention edges between them \citep{geva2023dissecting}. Specifically, we modify the additive mask $\mathbf{M}^{(l)}$ before the softmax to ablate a path $r$ as follows:
\begin{equation}
    \hat{M}_{ij}^{(l)} = M_{ij}^{(l)} - \infty \cdot \mathbf{1}_{(i,j) \in r}
\end{equation}
After the softmax, blocked positions have zero attention weight, and the remaining weights are automatically renormalized. Knockouts allow us to observe the impact of blocking attention to regions of the input tokens, to determine if models can solve a task without specific information.

\begin{figure}[t]
  \centering
  \includegraphics[width=1\textwidth]{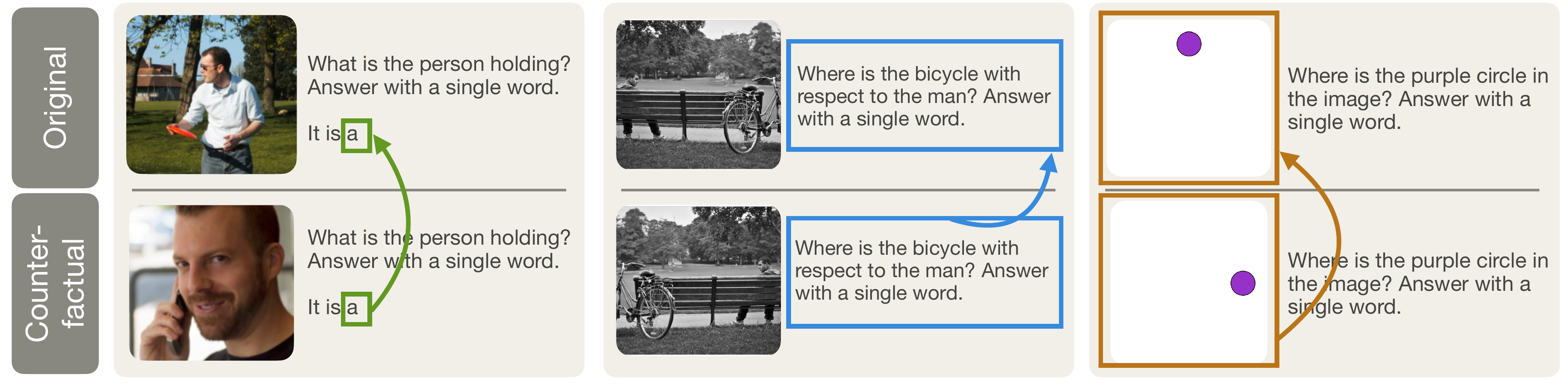}
  \caption{Causal patching interventions. We construct paired examples consisting of an original and a counterfactual sample with the same query but different visual content and answers. For each pair, we perform three independent patching interventions on the \hl{last_token_color}{last token}, \hl{image_patch_color}{text tokens}, or \hl{text_patch_color}{image token} representations. In each case, hidden representations from the counterfactual example are inserted into the forward pass of the original at a given layer, to measure from which representations the last token extracts information for its final prediction.}
  \label{fig:patching_setup}
\end{figure}

\paragraph{Intervention: Causal Patching.}

One limitation of attention knockout is that it measures model degradation when ablating a path, but not whether that pathway is used during normal inference. In particular, a minor degradation may indicate that the ablated pathway is not critical \textit{or} that alternative pathways can compensate. Thus, the majority of our analyses are based on interchange interventions \citep{geiger2025causal}. Let $\mathbf{h}_{\ell}(x_k)$ denote the hidden representation at layer $\ell$ of the token at position $k$. Given an original input $x$ and a counterfactual input $\tilde{x}$, we perform causal patching by feeding $x$ into the model and, at layer $\ell$, replacing the hidden representations with those from the corresponding tokens in the inference of $\tilde{x}$. For instance, for causal patching of image tokens at layer $\ell$ we replace the hidden representations of every token as follows:
\begin{equation}
\mathbf{h}'_{\ell}(x_k) = \begin{cases} \mathbf{h}_{\ell}(\tilde{x}_k) , & x_k \in \mathcal{I}, \\\ \mathbf{h}_{\ell}(x_k), & x_k \notin \mathcal{I}, \end{cases}
\end{equation}
Inference then proceeds normally. Text-token patching is defined analogously using $\mathcal{T}$. We measure whether the patched representation $\mathbf{h}'_{\ell}(x)$ contains sufficient information to shift the prediction toward the counterfactual output $\tilde{y}$ \citep{meng2022locating}, measured as the normalized increase in its probability. 

\begin{equation}
    \text{Restoration Score} = \frac{P(\tilde{y} \, | \,x \,;  \mathbf{h}'_{\ell}(x)) - P(\tilde{y} | x)}{P(\tilde{y}|\tilde{x}) - P(\tilde{y} | x)}
    \label{eq:causal_score}
\end{equation}
We only consider pairs where the model answers both $x$ and $\tilde{x}$ correctly under clean inference, and the restoration score is clipped to the range $[0,1]$, with values within this range left unchanged (values outside this range may occasionally occur due to numerical instability, such as when the difference between the clean and corrupted probabilities is extremely small). The exact input pairs (original and counterfactual), and patched components vary by experiments and are described in their respective sections, each chosen to isolate the causal role of a set of tokens. 

\begin{wraptable}{r}{0.55\textwidth}

    \centering
    \caption{Qwen3-VL-4B accuracy in synthetic and natural datasets under open and choices prompting schemes, including an ablation of visual input.}
    \label{tab:baseline_performance}
    \resizebox{\linewidth}{!}{%
    \begin{tabular}{llrrrr}
    \toprule
    Source & Task & $N$ & Choices & Open & NoImg \\
    \midrule
    \multirow{3}{*}{Synthetic} & Recognition & 400  & 100.0 & 100.0 & 24.5 \\
                             & Relations & 400  & 100.0 & 100.0 & 25.0 \\
                             & Localization & 400  & 100.0 & 100.0 & 24.0 \\
    \midrule
    \multirow{3}{*}{Natural}   & Recognition & 146  & 100.0   & 97.9  & 9.6  \\
                             & Relations & 1846 & 85.3  & 85.1  & 26.4 \\
                             & Localization & 3407 & 72.1  & 55.7  & 22.7 \\
    \bottomrule
    \end{tabular}
    }
\end{wraptable}
\paragraph{Models.} 
We study the behaviour of Qwen3-VL-4B, InternVL3.5-4B, and LLaVA-1.5-7B, covering different model families and training methodologies. In the main paper we present results for Qwen3-VL-4B, given its strong performance on the studied tasks. We replicate the key findings across model families and scales in Appendix \ref{sec:app_model_generalization}, including LLaVA-1.5, InternVL3.5, and Qwen3-VL-8B/32B. These experiments show that our findings generalize across model scales and architectures, suggesting that they are not an artifact of a single model or size. Table \ref{tab:baseline_performance} reports Qwen3-VL-4B performance on our tasks and datasets and confirms that visual information is necessary to solve them.

\section{Direct and text-mediated pathways}
\label{sec:task_dependence}

We aim to understand how information flows from the image to the final query token, whether it is read directly, routed through the text tokens, or both. To analyze this, we use causal patching on examples such that the original and counterfactual images yield different answers, while the text input remains identical (see Figure~\ref{fig:patching_setup}). We perform three independent patching interventions on the hidden states: (1) all image tokens, (2) all text query tokens, and (3) the final token. This allows us to identify when and which patched tokens change the output, i.e., which tokens carry the visual information necessary to solve the task. For the synthetic datasets, we construct the counterfactual pairs by changing the object, relation, or localization in each sample. For the natural datasets, for recognition we generate counterfactuals using the COCO recognition annotations of \citet{neo2024towards}, pairing samples with the same text query but using images of different objects. For the spatial relations and localization tasks, we filter left/right samples and horizontally flip the image (details in Appendix~\ref{sec:app_hflips}).

\begin{wraptable}{r}{0.55\textwidth}
\vspace{-4mm}
\centering
\small
\caption{Percentage of predictions aligned with text or image information when all text tokens are patched in all layers. For Relations the model relies on textual representations, whereas for Recognition it uses the image. Localization flips from image-dominated on synthetic data to mostly text-dominated on natural data.}
\label{tab:text_image_dominance}
\resizebox{\linewidth}{!}{
\begin{tabular}{llrrr}
\toprule
Source & Task & $N$ & Text (\%) & Image (\%)\\
\midrule
\multirow{3}{*}{Synthetic} & Recognition  & 200  & 0.0  & 100.0\\
                           & Relations   & 200  & 98.0 & 2.0  \\
                           & Localization & 200  & 0.0  & 100.0\\
\midrule
\multirow{3}{*}{Natural}   & Recognition  & 61   & 3.3  & 96.7 \\
                           & Relations   & 959  & 91.4 & 8.6  \\
                           & Localization & 1770 & 89.9 & 10.1 \\
\bottomrule
\end{tabular}
}

\end{wraptable}

\begin{figure}[t]
  \centering
  \includegraphics[width=1\textwidth]{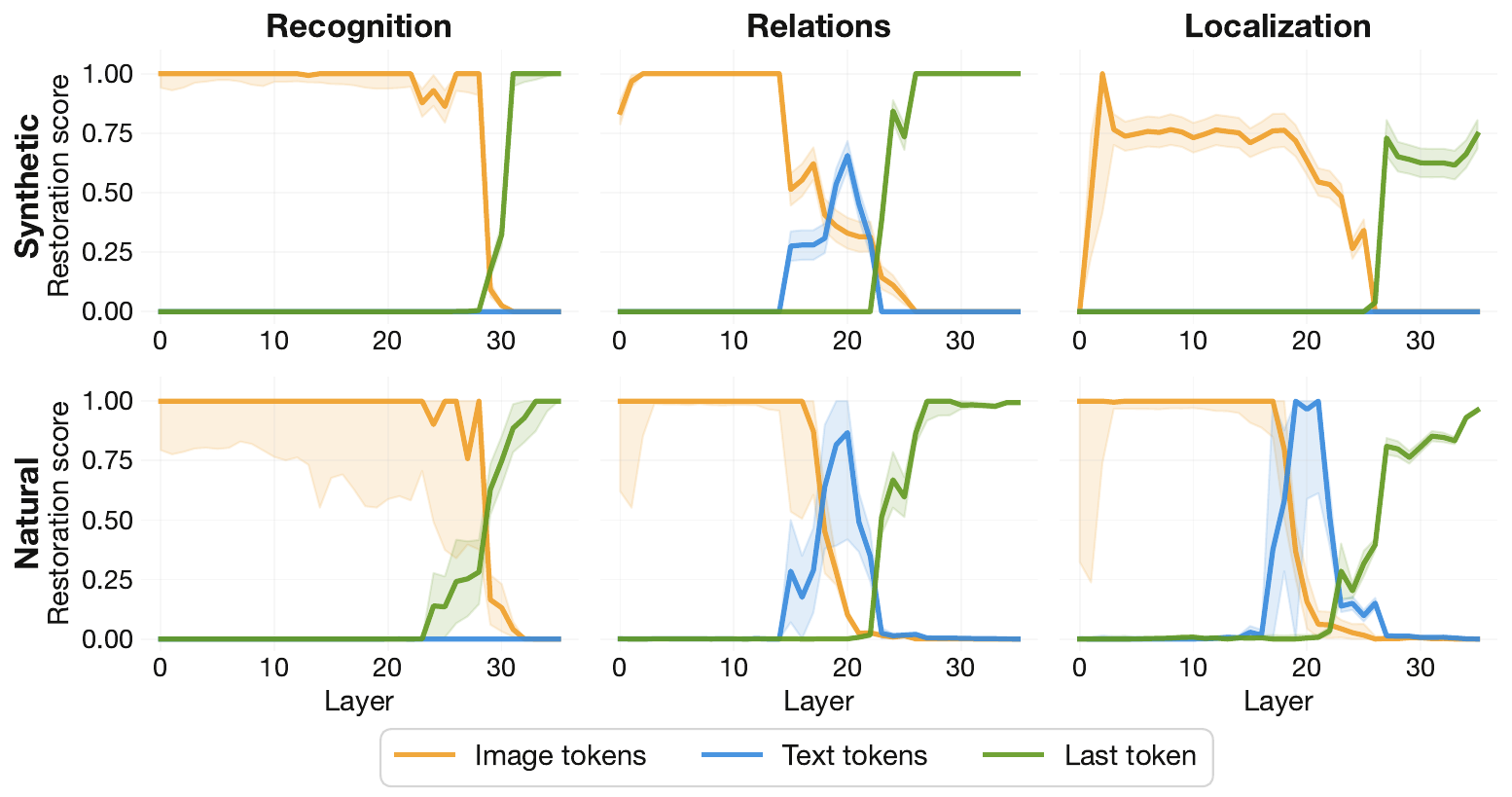}
  \caption{Restoration score of Qwen3-VL-4B under causal patching performed independently at each layer for different token groups. In recognition, the model uses the direct pathway, as textual patching has no effect at any layer. In spatial relation tasks, the text-mediated pathway is used, where the last token reads visual information from textual tokens. In localization, routing depends on the data distribution, with a direct pathway for synthetic data and a text-mediated pathway for natural data.}
  \label{fig:causal_patching}
\end{figure}

We first perform the patching interventions across \textit{all} layers (Table~\ref{tab:text_image_dominance}) to obtain a coarse measure of whether the model relies on information from the image or the text. In this setting,  image and text tokens provide opposing information. Then, we patch each layer independently (Figure~\ref{fig:causal_patching}) and measure the restoration score to reveal how information flows through the model and when cross-modal information transfer contributes to the final prediction. We additionally evaluate window patching (Appendix~\ref{sec:app_window_patching}) to account for information being shared across neighboring layers. We observe no difference from independent layer patching.

\paragraph{Object recognition uses the direct pathway.}
Table~\ref{tab:text_image_dominance} shows that when all layers are patched, predictions are aligned with the information from the image (100\% on synthetic and 97\% on natural data), ignoring the visual information encoded in the patched text representations. 
Figure~\ref{fig:causal_patching} reveals that this behavior is consistent when patching individual layers: restoration from image tokens remains high throughout the network, while patching text tokens has no effect at any layer. This indicates that object identity is encoded and preserved in the visual stream and is directly read out by the final token at later layers, without being transferred to the text tokens. These results show that object recognition is solved through a direct pathway, with no evidence of intermediate text mediation under standard inference.

\paragraph{Spatial relations are resolved through the text-mediated pathway.}
Table~\ref{tab:text_image_dominance} shows that spatial relation predictions are text-dominated when all layers are patched (98\% on synthetic and 91\% on natural data), indicating that the model relies on text representations to produce its final answer.
Patching individual layers (Figure~\ref{fig:causal_patching}) shows how the contributions of image and text representations to the prediction change with model depth. The visual information, initially encoded in the image tokens, is transferred to the text tokens at intermediate layers. After this transfer, patching the image tokens no longer affects the model prediction, indicating that the model relies on text representations. These results show that spatial relations are solved through text-mediated pathway, where visual information is routed through the text stream before being read out.

\paragraph{Localization uses either pathway depending on the data.}
For the localization task, models use either pathway depending on the data source. When patching the text tokens in all layers (Table~\ref{tab:text_image_dominance}), the model always predicts the answer using the image information on synthetic examples.\begin{wrapfigure}{r}{0.38\textwidth}
\vspace{-14pt}
  \centering
  \includegraphics[width=\linewidth]{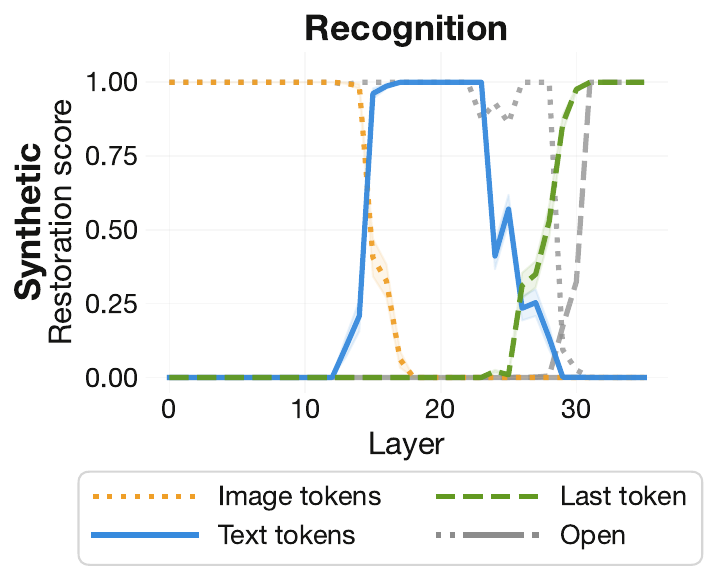}
  \caption{Prompt format modulates pathway selection. Causal patching on object recognition comparing generation protocols, \textit{choices} (colored) and \textit{open} (gray; Figure \ref{fig:causal_patching}, top left), shows that open prompts rely on the direct pathway, while choices induce text-mediated routing.}
  \label{fig:prompt_format}
\vspace{-40pt}
\end{wrapfigure} In contrast, on natural data, the model relies on the text representations (89.9\%) to predict the final answer, suggesting a shift towards the text-mediated pathway. The layer-by-layer results (Figure~\ref{fig:causal_patching}) confirm this contrast: text patching causes no restoration on synthetic data, whereas on natural datasets text reaches nearly 100\% of restoration in middle layers. This difference may reflect variation in visual complexity across datasets. Natural scenes often contain distractors, potentially requiring the model to first ground the queried object before localizing it.

This provides direct evidence that pathway selection is not fixed by the task itself: both routing strategies are available, and the input distribution can determine which one is used. If changes in the image source can induce this switch, a natural next question is whether changes in the text can induce a similar shift.

\paragraph{Prompt design can modulate pathway selection.}
We now analyze whether changes to the textual input affect the selected pathway. In particular, we compare the pathways used under the \textit{choices} and \textit{open} instruction protocols (\S\ref{sec:experimental_setup}). We apply paired causal patching to each protocol and layer independently and report restoration scores (Figure~\ref{fig:prompt_format}).

Since accuracy remains comparable across generation protocols (Table~\ref{tab:baseline_performance}), the differences in restoration reflect changes in pathway usage rather than task difficulty. We find that prompt design induces a clear within-task shift toward mediated routing for object recognition. As shown before, under the \textit{open} protocol, recognition relies on the direct pathway. In contrast, under the \textit{choices} protocol, text restoration becomes substantial across multiple layers, indicating a shift toward text-mediated routing (Figure \ref{fig:prompt_format}). For the other tasks, prompt design leads to small variations in restoration scores but does not change the dominant pathway: relations remain text-mediated, while localization is direct for synthetic data and mediated for natural data (Appendix~\ref{sec:app_prompt_modulation}).

However, prompt paraphrases that preserve the task and output format o not change the selected pathway (Appendix~\ref{sec:app_prompt_paraphrasing}), suggesting that pathway shifts are driven by changes in task requirements rather than wording alone.

\FloatBarrier
\section{Text mediation as a fallback pathway}
We characterized in \S\ref{sec:task_dependence} which routes VLMs use under normal inference: direct pathway for recognition, text-mediated pathway for spatial relations, and either pathway for localization, depending on data type. We now ask whether the unused route is genuinely inactive or remains latently available. This matters for two reasons. First, it enables a more complete characterization of visual information flow. Second, it highlights limitations in how ablation-based interventions should be interpreted. To address these questions, we use two approaches: (1) we show that text tokens alone are sufficient by replacing the input image with uniform noise and patching text representations from a counterfactual example containing the original image; and (2) we show that the pathway the model selects on standard inference is not the only viable route, by knocking out attention along each pathway and observing that the model can still solve the tasks. This helps reconcile conflicting conclusions in prior work, where the same or similar tasks are described as either direct or text-mediated depending on the intervention \citep{neo2024towards, zhang2025cross, kaduri2025s, kim2025interpreting, serra2025narrow, takishita2025llms}.

\subsection{Image corruption reveals that text tokens encode enough visual signal}\label{subsec:corrupted_patching}

We test whether text tokens alone carry sufficient visual information to recover the correct answer. To do so, we use corrupted patching, where we input a uniform noise image and patch, across all layers, the text representations from an inference with the original image (Figure~\ref{fig:corrupted_setup}). We also evaluate alternative image ablations using black and white images in Appendix~\ref{sec:app_corruption_ablation}.

\begin{figure}[t] 
    \begin{minipage}[t]{0.42\textwidth}
        \centering
        \vspace{0pt} 
        \includegraphics[width=0.85\linewidth]{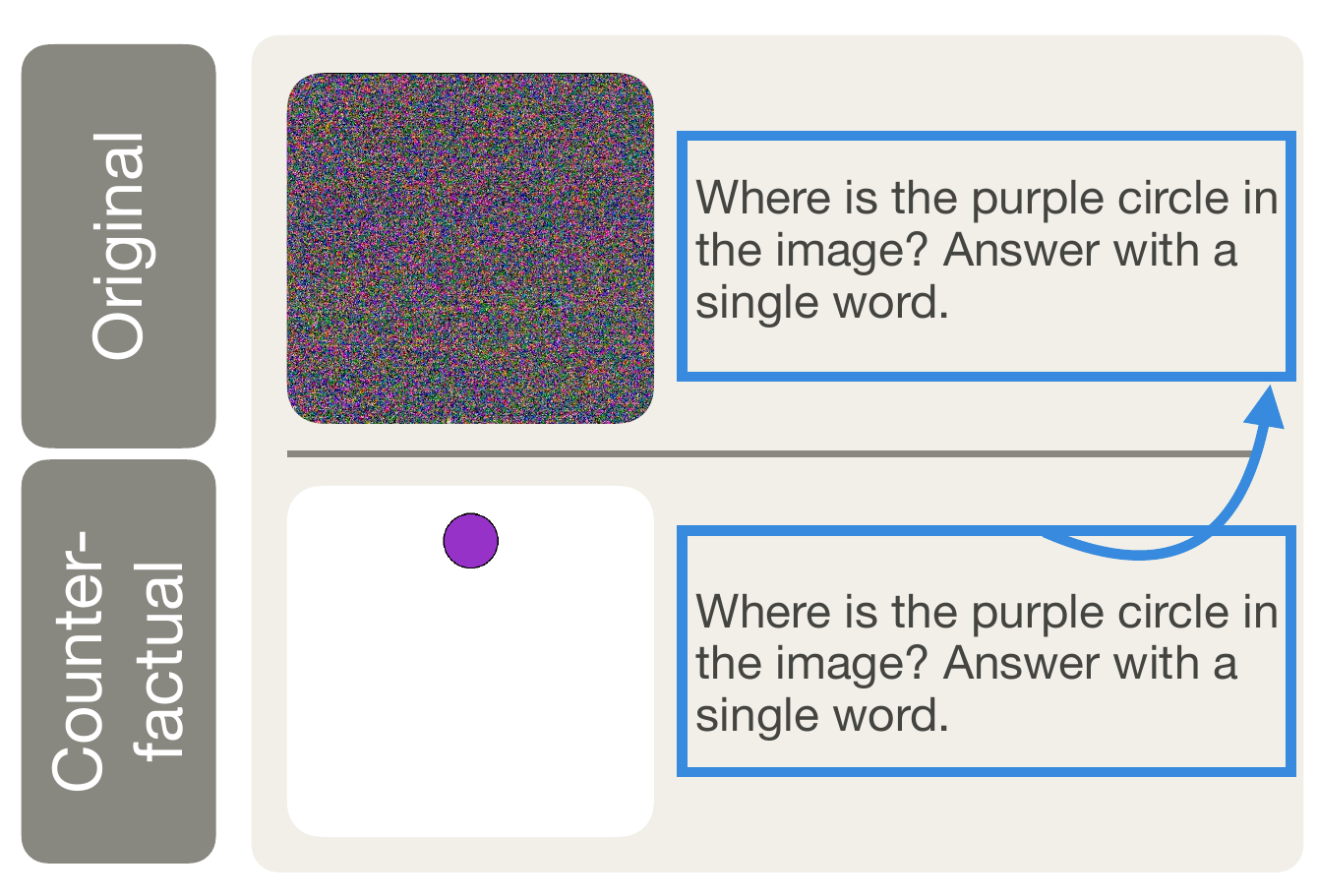}
        \captionof{figure}{Corrupted patching. Text hidden representations are inserted in the inference that contains a random noise image.}
        \label{fig:corrupted_setup}
    \end{minipage}%
    \hfill 
    \begin{minipage}[t]{0.56\textwidth}
        \centering
        \vspace{0pt} 
        \captionof{table}{Corrupted patching results. RS denotes the restoration score. The model recovers the correct answer using only the text-mediated route, even in tasks where the direct pathway dominates under normal inference.}
        \label{tab:text_recovery_results}
        \resizebox{\linewidth}{!}{%
            \begin{tabular}{ll ccc ccc}
            \toprule
            \multirow{2}{*}{\textbf{Source}} & \multirow{2}{*}{\textbf{Task}} & \multicolumn{3}{c}{\textbf{Open}} & \multicolumn{3}{c}{\textbf{Choices}} \\
            \cmidrule(lr){3-5} \cmidrule(lr){6-8}
            & & $N$ & RS & Acc (\%) & $N$ & RS & Acc (\%) \\
            \midrule
            \multirow{3}{*}{Synth.} & Recog.  & 400 & 1.00 & 100.0 & 400 & 1.00 & 100.0 \\
                                     & Relat.  & 400 & 1.00 & 100.0 & 400 & 1.00 & 100.0 \\
                                     & Local.  & 400 & 0.08 & 100.0 & 400 & 0.80 & 100.0 \\
            \midrule
            \multirow{3}{*}{Nat.}   & Recog.  & 144 & 0.82 & 81.9  & 146 & 0.99 & 99.3  \\
                                     & Relat.  & 1361& 1.00 & 94.9  & 1401& 0.98 & 97.9  \\
                                     & Local.  & 1897& 1.00 & 95.8  & 2471& 0.92 & 92.7  \\
            \bottomrule
            \end{tabular}%
        }
    \end{minipage}
\end{figure} %

Table~\ref{tab:text_recovery_results} shows the restoration score and accuracy for corrupted patching. We see that across tasks and datasets, accuracy is high, with the lowest value being 81.9\% for natural-image recognition. In terms of restoration score, spatial relations and natural-image localization achieve a score of 1.0, consistent with their reliance on text-mediated routing. More strikingly, we see the same result for object recognition, on both synthetic and natural datasets, despite this task relying on the direct pathway under normal inference. This indicates that the model can recover the correct answer using only the patched, visually-informed, text-token representations, showing that the text tokens encode sufficient visual information, even for tasks where the model usually relies on the direct pathway.

On the other hand, synthetic localization obtains 100\% accuracy but near 0 restoration score. We note that this difference happens because the restoration score is computed from unrestricted next-token probabilities, while accuracy is the top1 token from the restricted candidate set (Section \ref{sec:experimental_setup}). Looking at the outputs of the corrupted patched inference, the model generates spatially relevant responses outside the candidate set used for evaluation (e.g., ``center'' instead of \{top, bottom, left, right\}), resulting in a low probability for the correct answer and thus a low restoration score. Nevertheless, the correct spatial position remains the highest-ranked candidate in all patched samples (Appendix~\ref{sec:app_synthetic_loc_outputs}), demonstrating that the patched text representations retain sufficient spatial information for the model to identify the correct position among the candidate answers.

The apparent discrepancy of corrupted patching results with the causal patching results (\S\ref{sec:task_dependence}) is informative. Under paired causal patching, text tokens do not override the image signal because both pathways carry competing information. In contrast, under corrupted inputs, the image provides no competing signal, allowing the text trajectory to fully determine the prediction. These results show that text tokens consistently encode sufficient visual information to solve the task, but under standard inference they act as a secondary source when the direct visual pathway is available. When the direct pathway is removed, the model can rely entirely on the text-mediated route.

\subsection{Attention knockouts reveal pathway rerouting}

The previous experiment showed that text tokens contain sufficient visual information to solve the task. We now ask whether either pathway is necessary. To answer this we use attention knockouts, selectively blocking attention along the direct pathway ($r_{\mathcal{I} \to \text{last}}$) and the mediated pathway ($r_{\mathcal{I} \to \mathcal{T}}$, $r_{\mathcal{T} \to \text{last}}$) at all layers. We measure the resulting change in accuracy.

\begin{wraptable}{r}{0.55\textwidth}
\vspace{-12pt}
\centering
\caption{Accuracy changes after attention knockout on Qwen3-VL-4B. Blocking the direct pathway ($r_{\mathcal{I} \to \text{last}}$) has negligible effect across tasks, while disrupting the text-mediated pathway, causes substantial drops, indicating that the direct pathway is not necessary for task performance.}
\label{tab:knockout_summary}
\resizebox{\linewidth}{!}{
\begin{tabular}{llrrr}
\toprule
Source & Task & $r_{\mathcal{I} \to \text{last}}$ & $r_{\mathcal{I} \to \mathcal{T}}$ & $r_{\mathcal{T} \to \text{last}}$ \\
\midrule
\multirow{3}{*}{Synthetic} & Recognition  & 0.0 & 0.0  & 0.0 \\
                           & Relations    & 0.0 & -10.5 & 0.0 \\
                           & Localization & 0.0 & 0.0  & 0.0 \\
\midrule
\multirow{3}{*}{Natural}   & Recognition  & +1.4 & -2.7  & +2.1 \\
                           & Relations    & -1.5 & -34.9 & -1.5 \\
                           & Localization & +3.6 & -18.3 & -1.6 \\
\bottomrule
\end{tabular}
}
\vspace{-15pt}
\end{wraptable}

Blocking the direct pathway ($r_{\mathcal{I} \to \text{last}}$) has almost no effect on performance, as shown in Table~\ref{tab:knockout_summary}. Combined with the previous corrupted-patching experiment, this shows that the model can reroute through the text-mediated pathway when the direct one is ablated.

In contrast, blocking the first stage of the text-mediated pathway ($r_{\mathcal{I} \to \mathcal{T}}$) shows the biggest impact on performance. The magnitude of these drops is consistent with the preferred pathway identified by causal patching (\S\ref{sec:task_dependence}), with tasks relying on text-mediation being affected the most. 

Blocking the second stage ($r_{\mathcal{T} \to \text{last}}$) causes much smaller drops, possibly because visual information could flow through special tokens in the prompt (which are not blocked in this setup). Although, it could also mean that the model is rerouting through the direct pathway, we leave this distinction for future work. 

Recent observations have shown that the final token places limited attention on image tokens~\citep{chen2025spatial, dalal2025constructive}, positioning this as a problem to be solved. Table \ref{tab:knockout_summary} shows that the final token does not need to attend to the image. The model can rely on visual information already encoded in the text tokens. Moreover, correction interventions that modify the attention distribution of the last token by boosting attention to image tokens only focus on the direct pathway. Since several tasks are solved through text-mediation, our results suggest that intervening on image-to-query or query-to-last interactions may be a promising direction.

\FloatBarrier
\section{Related Work}
\paragraph{Information flow.} Some prior work focuses on studying how visual representations are transformed through the network \citep{fu2025hidden, liu2025visual}. Other work analyzes specific subsets of tokens, such as high-norm or sink tokens, and their role in preserving semantic information \citep{luo2025sink, basu2024understanding}. More closely related to our work, a second line of research investigates how visual information reaches the final prediction. Using attention knockouts, \citet{kaduri2025s} show that text tokens encode global visual information that enables image descriptions, while \citet{zhang2025cross} focus on image-to-text information transfer and concludes that all tasks rely on text-mediated pathways. Focusing on object-related information flow, \citep{neo2024towards} show with knockout experiments that blocking the last token from accessing the image drops performance, suggesting a direct-readout mechanism. These studies reach different conclusions regarding whether final predictions rely on text-mediated pathways or direct access to image representations, whereas we show that both pathways coexist within the same model.

\paragraph{Task-specific mechanistic interpretability.} Several recent studies investigate the mechanisms used by VLMs when solving specific tasks. \citet{kang2026linear} show that object-specific \textit{text} tokens encode spatial information, supporting a text-mediated pathway, whereas \citet{cui2026dual} find that the final token can directly retrieve attributes from image representations, suggesting a direct readout mechanism. We suggest these differences reflect task and dataset structure rather than fundamentally distinct mechanisms. In particular, \citet{kang2026linear} study a natural dataset on what we refer to here as a relational task (``Is the dog to the left or right of the cat?''), which requires comparing the positions of two objects, whereas \citet{cui2026dual} experiment on controlled images on a \textit{recognition} task (e.g ``What is the colour of the square to the left of the green square?''). Beyond spatial reasoning, work on factual recall and object identification has shown that image representations are transformed within the VLM and carry task-relevant information that is later accessed by the last token \citep{basu2024understanding, venhoff2025visual, li2026causal}. We show that the  mechanism use by the model depends on the task being solved.

\paragraph{Interventions for understanding and controlling VLMs.} A variety of interventions have been proposed to analyze and modify visual information flow in VLMs. Attention knockouts and attention patterns are often used to motivate and design test-time interventions. \citet{chen2025spatial} proposes modifying attention from the last token to image tokens, while \citet{dalal2025constructive} uses attention patterns from generated answer tokens to modify the input image before generating a new answer. In contrast, \citet{li2026causal} uses activation patching with corrupted images to identify where to add residual connections across layers for the last token at test time. Our results suggest that, for tasks relying on text mediation, increasing attention to query tokens or leveraging their attention patterns rather than only those of answer tokens may lead to more effective interventions. Our findings also suggest caution when designing interventions based on corrupted-image patching, as corruption may induce mechanisms different from those used under normal conditions.

\section{Discussion}
\label{sec:discussion}
\paragraph{Interpreting pathway usage in VLMs.}
Our results highlight an interpretational gap in ablation-based interpretability methods. Attention knockouts answer a necessity question (whether a model can solve a task without a given component), while causal patching addresses a usage question (whether the model reads information from that component in its standard forward computation). These notions can diverge, for instance, ablating the direct pathway does not affect recognition accuracy, despite causal patching showing that recognition relies entirely on the direct pathway. 

Instead, we argue that characterizing a pathway requires combining analyses, paired causal patching identifies the pathway used in the model’s standard forward computation, attention knockouts test whether it is necessary, and corrupted patching tests whether the alternative text-mediated pathway is sufficient. Relying on any single method can lead to incomplete or misleading conclusions. %

This pattern extends observations from the backup-circuit literature in text-only models~\citep{wang2023interpretability, mcgrath2023hydra, mcdougall-etal-2024-copy}. While prior work showed partial compensation when components are ablated, we observe almost full rerouting in VLMs, with models switching pathways and maintaining task performance rather than degrading. Default mechanisms in VLMs are therefore not reliably captured by ablation-based interventions alone, and instead require usage-sensitive interventions in order to be identified.

\paragraph{Implications for VLM design and reasoning.}
First, the direct pathway is not necessary for any evaluated task: text mediation alone recovers correct answers even when the image is removed and, in several cases, ablating the direct pathway slightly improves performance. This suggests that the direct pathway can sometimes act as a noisy competing signal rather than a necessary information route. Whether this observation could motivate to test-time interventions or training-time architectural choices remains an open question for future work. 

Second, our results suggest a functional distinction between pathways. Text mediation dominates compositional tasks such as spatial relations and becomes more prominent with visual and prompt complexity. One interpretation is that image tokens primarily extract attributes, while the bulk of computation, reference resolution, integration, and combination of alternatives, happens in the text-token stream; the direct pathway serves as a fast route when attribute lookup suffices. Whether these text-stream operations constitute genuine reasoning or structured feature aggregation is an open question requiring further analysis.

\paragraph{Limitations.}
Our experiments cover three task types (recognition, localization, spatial relations) evaluated on three open-weight VLM families as single-token predictions; other tasks (e.g., captioning, multi-step visual reasoning) and larger models may engage mechanisms we do not characterize. Our pathway analysis operates at the level of token representation groups, so identifying the specific attention heads and MLP components that implement each pathway is left to future work. Finally, our analysis is inference-time only and does not address how these pathways emerge during training.

\section{Conclusion}
In this work, we study and characterize how visual information is integrated in VLMs during text generation. We find that VLMs route visual information through two pathways: a \textit{direct pathway}, where the last token reads visual information directly from image tokens, and a \textit{text-mediated pathway}, where visual information is first transferred to query tokens and later read out by the last token. We also find that pathway selection is task-dependent and modulated by data distribution and prompt design. Finally, we show that text-mediated routing remains available even on tasks normally solved through direct readout, with text representations encoding sufficient information to recover the answer, suggesting that interventions can activate alternative routing strategies in VLMs. More broadly, our findings show that visual information can be integrated through both image and text representations, highlighting the importance of studying and intervening on both direct visual readout and text-mediated transfer of visual information.

\section*{Acknowledgments}
This work was supported by research grant (VIL53122) from Villum Fonden. SF was supported by NNF project 0094281 from Novo Nordisk Fonden. DO was supported by a grant of the Ministry of Research, Innovation and Digitization, CNCS-UEFISCDI, project number PN-IV-P2-2.1-TE-2023-1632 within PNCDI IV. CF was supported by Danish Data Science Academy, which is funded by the Novo Nordisk Foundation (NNF21SA0069429). We acknowledge the EuroHPC Joint Undertaking for awarding this project access to the EuroHPC supercomputer LEONARDO, hosted by CINECA (Italy) and the LEONARDO consortium through an EuroHPC Development Access call (ID:EUHPC\_D27\_102).

\bibliographystyle{plainnat}
\bibliography{custom}

@string{aaai = {Proc. AAAI}}

@string{colm = {Proc. COLM}}

@string{cvpr = {Proc. CVPR}}

@string{eccv = {Proc. ECCV}}

@string{emnlp = {Proc. EMNLP}}

@string{icml = {Proc. ICML}}

@string{iclr = {Proc. ICLR}}

@string{ijcv = {Int. J. Comput. Vis.}}

@string{jmlr = {J. Mach. Learn. Res.}}

@string{neurips = {Proc. NeurIPS}}

@string{tacl = {Trans. Assoc. Comput. Linguistics}}

@string{tmlr = {Trans. Mach. Learn. Res.}}

@inproceedings{kamath2023s,
    title={What's “up” with vision-language models? {I}nvestigating their struggle with spatial reasoning},
    author={Kamath, Amita and Hessel, Jack and Chang, Kai-Wei},
    booktitle=emnlp,
    year={2023}
}

@inproceedings{chen2025spatial,
    title={Why is spatial reasoning hard for {VLM}s? {A}n attention mechanism perspective on focus areas},
    author={Chen, Shiqi and Zhu, Tongyao and Zhou, Ruochen and Zhang, Jinghan and Gao, Siyang and Niebles, Juan Carlos and Geva, Mor and He, Junxian and Wu, Jiajun and Li, Manling},
    booktitle=icml,
    year={2025}
}

@inproceedings{neo2024towards,
  title={Towards interpreting visual information processing in vision-language models},
  author={Neo, Clement and Ong, Luke and Torr, Philip and Geva, Mor and Krueger, David and Barez, Fazl},
  booktitle=iclr,
  year={2025}
}

@inproceedings{meng2022locating,
  title={Locating and editing factual associations in {GPT}},
  author={Meng, Kevin and Bau, David and Andonian, Alex and Belinkov, Yonatan},
  booktitle=neurips,
  year={2022}
}

@inproceedings{li2026causal,
  title={Causal tracing of object representations in large vision language models: Mechanistic interpretability and hallucination mitigation},
  author={Li, Qiming and Ye, Zekai and Feng, Xiaocheng and Zhong, Weihong and Ma, Weitao and Feng, Xiachong},
  booktitle=aaai,
  year={2026}
}

@article{liu2023visual,
  author       = {Fangyu Liu and Guy Emerson and Nigel Collier},
  title        = {Visual Spatial Reasoning},
  journal      = tacl,
  volume       = {11},
  pages        = {635--651},
  year         = {2023},
}

@inproceedings{kang2026linear,
  title={Linear Mechanisms for Spatiotemporal Reasoning in Vision Language Models},
  author={Kang, Raphi and Chen, Hongqiao and Gkioxari, Georgia and Perona, Pietro},
  booktitle=iclr,
  year={2026}
}

@inproceedings{bai2025qwen3,
  title={Qwen3-{VL} technical report},
  author={Bai, Shuai and Cai, Yuxuan and Chen, Ruizhe and Chen, Keqin and Chen, Xionghui and Cheng, Zesen and Deng, Lianghao and Ding, Wei and Gao, Chang and Ge, Chunjiang and others},
  journal={arXiv preprint arXiv:2511.21631},
  year={2025}
}

@inproceedings{zhang2025cross,
  title={Cross-modal information flow in multimodal large language models},
  author={Zhang, Zhi and Yadav, Srishti and Han, Fengze and Shutova, Ekaterina},
  booktitle=cvpr,
  year={2025}
}

@article{kim2025interpreting,
  title={Interpreting Attention Heads for Image-to-Text Information Flow in Large Vision-Language Models},
  author={Kim, Jinyeong and Kang, Seil and Park, Jiwoo and Kim, Junhyeok and Hwang, Seong Jae},
  journal={arXiv preprint arXiv:2509.17588},
  year={2025}
}

@inproceedings{kaduri2025s,
  title={What's in the Image? {A} Deep-Dive into the Vision of Vision Language Models},
  author={Kaduri, Omri and Bagon, Shai and Dekel, Tali},
  booktitle=cvpr,
  year={2025}
}

@inproceedings{serra2025narrow,
  title={The narrow gate: Localized image-text communication in native multimodal models},
  author={Serra, Alessandro Pietro and Ortu, Francesco and Panizon, Emanuele and Valeriani, Lucrezia and Basile, Lorenzo and Ansuini, Alessio and Doimo, Diego and Cazzaniga, Alberto},
  booktitle=neurips,
  year={2025}
}

@inproceedings{lin2014microsoft,
  title={Microsoft {COCO}: Common objects in context},
  author={Lin, Tsung-Yi and Maire, Michael and Belongie, Serge and Hays, James and Perona, Pietro and Ramanan, Deva and Doll{\'a}r, Piotr and Zitnick, C Lawrence},
  booktitle=eccv,
  year={2014},
}

@article{krishna2017visual,
  title={Visual genome: Connecting language and vision using crowdsourced dense image annotations},
  author={Krishna, Ranjay and Zhu, Yuke and Groth, Oliver and Johnson, Justin and Hata, Kenji and Kravitz, Joshua and Chen, Stephanie and Kalantidis, Yannis and Li, Li-Jia and Shamma, David A and others},
  journal=ijcv,
  volume={123},
  number={1},
  pages={32--73},
  year={2017},
  publisher={Springer}
}

@article{geiger2025causal,
  title={Causal abstraction: A theoretical foundation for mechanistic interpretability},
  author={Geiger, Atticus and Ibeling, Duligur and Zur, Amir and Chaudhary, Maheep and Chauhan, Sonakshi and Huang, Jing and Arora, Aryaman and Wu, Zhengxuan and Goodman, Noah and Potts, Christopher and others},
  journal=jmlr,
  volume={26},
  number={83},
  pages={1--64},
  year={2025}
}

@inproceedings{wang2023interpretability,
    title={Interpretability in the Wild: a Circuit for Indirect Object Identification in {GPT}-2 Small},
    author={Kevin Ro Wang and Alexandre Variengien and Arthur Conmy and Buck Shlegeris and Jacob Steinhardt},
    booktitle=iclr,
    year={2023},
    url={https://openreview.net/forum?id=NpsVSN6o4ul}
}

@article{mcgrath2023hydra,
  title={The hydra effect: Emergent self-repair in language model computations},
  author={McGrath, Thomas and Rahtz, Matthew and Kramar, Janos and Mikulik, Vladimir and Legg, Shane},
  journal={arXiv preprint arXiv:2307.15771},
  year={2023}
}

@inproceedings{mcdougall-etal-2024-copy,
    title = "Copy Suppression: Comprehensively Understanding a Motif in Language Model Attention Heads",
    author = "McDougall, Callum Stuart  and Conmy, Arthur  and Rushing, Cody  and McGrath, Thomas  and Nanda, Neel",
    booktitle = "Proceedings of the 7th BlackboxNLP Workshop: Analyzing and Interpreting Neural Networks for NLP",
    year = "2024",
    url = "https://aclanthology.org/2024.blackboxnlp-1.22/",
    doi = "10.18653/v1/2024.blackboxnlp-1.22",
}

@article{wang2025internvl3,
  title={Intern{VL}3.5: Advancing open-source multimodal models in versatility, reasoning, and efficiency},
  author={Wang, Weiyun and Gao, Zhangwei and Gu, Lixin and Pu, Hengjun and Cui, Long and Wei, Xingguang and Liu, Zhaoyang and Jing, Linglin and Ye, Shenglong and Shao, Jie and others},
  journal={arXiv preprint arXiv:2508.18265},
  year={2025}
}

@inproceedings{takishita2025llms,
  title={{LLM}s Can Compensate for Deficiencies in Visual Representations},
  author={Takishita, Sho and Gala, Jay and Mohamed, Abdelrahman and Inui, Kentaro and Kementchedjhieva, Yova},
  booktitle={EMNLP Findings},
  year={2025}
}

@article{fu2025hidden,
  title={Hidden in plain sight: {VLM}s overlook their visual representations},
  author={Fu, Stephanie and Bonnen, Tyler and Guillory, Devin and Darrell, Trevor},
  journal={arXiv preprint arXiv:2506.08008},
  year={2025}
}

@article{venhoff2025visual,
  title={How visual representations map to language feature space in multimodal llms},
  author={Venhoff, Constantin and Khakzar, Ashkan and Joseph, Sonia and Torr, Philip and Nanda, Neel},
  journal={arXiv preprint arXiv:2506.11976},
  year={2025}
}

@article{cui2026dual,
  title={The Dual Mechanisms of Spatial Reasoning in Vision-Language Models},
  author={Cui, Kelly and Prakash, Nikhil and Raina, Ayush and Bau, David and Torralba, Antonio and Shaham, Tamar Rott},
  journal={arXiv preprint arXiv:2603.22278},
  year={2026}
}

@inproceedings{liu2025visual,
  title={Visual representations inside the language model},
  author={Liu, Benlin and Kamath, Amita and Grunde-McLaughlin, Madeleine and Han, Winson and Krishna, Ranjay},
  booktitle=colm,
  year={2025}
}

@inproceedings{luo2025sink,
  title={To sink or not to sink: Visual information pathways in large vision-language models},
  author={Luo, Jiayun and Fan, Wan-Cyuan and Wang, Lyuyang and He, Xiangteng and Rahman, Tanzila and Abolmaesumi, Purang and Sigal, Leonid},
  booktitle=iclr,
  year={2026}
}

@inproceedings{basu2024understanding,
  title={Understanding information storage and transfer in multi-modal large language models},
  author={Basu, Samyadeep and Grayson, Martin and Morrison, Cecily and Nushi, Besmira and Feizi, Soheil and Massiceti, Daniela},
  booktitle=neurips,
  year={2024}
}

@inproceedings{
geva2023dissecting,
title={Dissecting Recall of Factual Associations in Auto-Regressive Language Models},
author={Mor Geva and Jasmijn Bastings and Katja Filippova and Amir Globerson},
booktitle=emnlp,
year={2023},
url={https://openreview.net/forum?id=F1G7y94K02}
}

@InProceedings{liu2024improved,
    author    = {Liu, Haotian and Li, Chunyuan and Li, Yuheng and Lee, Yong Jae},
    title     = {Improved Baselines with Visual Instruction Tuning},
    booktitle = cvpr,
    month     = {June},
    year      = {2024},
    pages     = {26296-26306}
}

@article{dalal2025constructive,
  title={Constructive distortion: Improving mllms with attention-guided image warping},
  author={Dalal, Dwip and Vashishtha, Gautam and Mishra, Utkarsh and Kim, Jeonghwan and Kanda, Madhav and Ha, Hyeonjeong and Lazebnik, Svetlana and Ji, Heng and Jain, Unnat},
  journal=iclr,
  year={2025}
}

@article{sharkey2025open,
  title={Open Problems in Mechanistic Interpretability},
  author={Sharkey, Lee and Chughtai, Bilal and Batson, Joshua and Lindsey, Jack and Wu, Jeff and Bushnaq, Lucius and Goldowsky-Dill, Nicholas and Heimersheim, Stefan and Ortega, Alejandro and Bloom, Joseph and others},
  journal=tmlr,
  year={2025}
}

\newpage{}
\appendix
\raggedbottom

\addcontentsline{toc}{section}{Appendix}
\section*{Appendix}
\startcontents[appendix]
\printcontents[appendix]{}{1}{\subsection*{Appendix Contents}}

\clearpage{}
\FloatBarrier
\section{Datasets and Experimental Setup}
\label{sec:app_datasets}

\FloatBarrier
\subsection{Synthetic Shapes Datasets}
\label{sec:app_datasets_synth}
We generate three synthetic datasets, \textit{Shapes Recognition}, \textit{Shapes Localization}, and \textit{Shapes Relations}, using a single pipeline. The shared generation ensures all three datasets are constructed from the same primitives and spatial grid, so that paired samples differ only in the feature targeted by each task. Examples are shown in Figure~\ref{fig:tasks_samples}.

\paragraph{Visual primitives.}
Each image is rendered on a white background with shapes drawn from a fixed inventory: 6 shape types (circle, square, triangle, star, diamond, pentagon) and 6 colors (red, blue, green, yellow, orange, purple). For \textit{Shapes Recognition}, we restrict to four shapes (circle, square, triangle, star) when constructing the candidate set under the \textit{choices} setting. Shapes are rendered with a black outline at a fixed size relative to the image. Each synthetic dataset contains 400 samples.

\paragraph{Spatial layout.}
Shapes are placed at the centers of cells of a $3 \times 3$ grid. The four cardinal cells (top, bottom, left, right of the center) are used as candidate positions; the center cell is reserved for the second object in \textit{Shapes Relations}. 
\begin{itemize} 
    \item \textit{Shapes Recognition}: a single shape is placed in one of the four cardinal cells. The task is to identify the shape.
    \item \textit{Shapes Localization}: identical layout to Recognition; the task is to identify the position (top, bottom, left, right) of the shape relative to the image center.
    \item \textit{Shapes Relations}: a second shape is added at the center. The task is to identify the spatial relation (left, right, above, below) of the off-center shape with respect to the centered one.
\end{itemize}

\paragraph{Counterfactual pairs.}
Because all three datasets share the same primitives and grid, paired samples can be constructed by varying only the feature relevant to the task:
\begin{itemize}
    \item \textit{Recognition}: the shape identity is changed while keeping color and position fixed.
    \item \textit{Localization}: the position is changed while keeping shape and color fixed.
    \item \textit{Relations}: the off-center shape is moved to a different cardinal cell, yielding a different relation while preserving the two objects.
\end{itemize}
This shared construction supports clean causal interventions, since within a pair the visual input differs only on the controlled axis.

\FloatBarrier
\subsection{Natural Datasets}
\label{sec:app_datasets_nat}

We evaluate on five natural-image sources covering the three task types. Sample sizes refer to the post-filtering subsets used in our experiments. Per-dataset baseline accuracy after filtering is reported in Table~\ref{tab:hflip_baselines} for the horizontal-flip variants. Examples per task are shown in Figure~\ref{fig:tasks_samples}.

\paragraph{Recognition.}
\begin{itemize}
    \item \textbf{COCO recognition}: COCO images paired with the recognition queries by~\citet{neo2024towards}.
\end{itemize}

\paragraph{Spatial relations.}
\begin{itemize}
    \item \textbf{Controlled CLEVR}: Two-object spatial-relation scenes from the controlled CLEVR split of the What's Up benchmark~\citep{kamath2023s}.
    \item \textbf{Controlled Images}: Two-object spatial-relation queries on natural photographs, from the controlled images split of What's Up~\citep{kamath2023s}.
    \item \textbf{COCO\_two}: Two-object spatial-relation queries based on COCO~\citep{lin2014microsoft} images, from the natural COCO split of What's Up.
    \item \textbf{VG\_QA\_two}: Two-object spatial-relation queries based on Visual Genome~\citep{krishna2017visual} images, from the natural VG split of What's Up.
    \item \textbf{VSR}: The Visual Spatial Reasoning benchmark~\citep{liu2023visual}, naturalistic spatial-relation queries spanning a broader set of relations. We adapt the queries to our evaluation protocol (see \S\ref{sec:app_prompts}).
\end{itemize}

\paragraph{Localization.}
\begin{itemize}
    \item \textbf{COCO\_one}: Single-object localization queries on COCO images, from the COCO single-object split of What's Up~\citep{kamath2023s, lin2014microsoft}.
    \item \textbf{VG\_QA\_one}: Single-object localization queries on Visual Genome images, from the VG single-object split of What's Up~\citep{kamath2023s, krishna2017visual}.
\end{itemize}

\paragraph{Counterfactual construction on natural data.}
For \emph{relations} and \emph{localization} on natural data, we construct counterfactuals by horizontally flipping the image and inverting the left/right ground-truth answer. We restrict to samples whose ground-truth answer involves a left/right relation and where the model produces the correct answer on both members of the pair under clean inference. Construction details and per-dataset filtered sizes are reported in \S\ref{sec:app_hflips}. For \emph{recognition}, counterfactual pairs share the same query but use images of different objects.

\paragraph{Sample sizes and per-dataset contribution to the natural evaluation.}
Table~\ref{tab:app_natural_sizes} reports the post-filtering sample count for each natural dataset, alongside its proportion within its task aggregate and within the full natural evaluation.

\begin{table}[htbp]
\centering
\caption{Natural dataset sizes. Post-filtering sample counts used as the natural evaluation set. \% Task is the dataset's share of its task aggregate; \% Natural is its share of the combined natural evaluation across all tasks.}
\label{tab:app_natural_sizes}
\small
\begin{tabular}{llrrr}
\toprule
Dataset & Task & N & \% Task & \% Natural \\
\midrule
Controlled CLEVR    & relations    & 408  & 22.1 & 7.6  \\
Controlled Images   & relations    & 412  & 22.3 & 7.6  \\
COCO\_two           & relations    & 440  & 23.8 & 8.2  \\
VG\_QA\_two         & relations    & 288  & 15.6 & 5.3  \\
VSR                 & relations    & 298  & 16.1 & 5.5  \\
\textit{Subtotal}   & \textit{relations}    & \textit{1{,}846} & \textit{100.0} & \textit{34.2} \\
\midrule
COCO\_one           & localization & 2{,}247 & 66.0 & 41.6 \\
VG\_QA\_one         & localization & 1{,}160 & 34.0 & 21.5 \\
\textit{Subtotal}   & \textit{localization} & \textit{3{,}407} & \textit{100.0} & \textit{63.1} \\
\midrule
COCO recognition    & recognition  & 146  & 100.0 & 2.7  \\
\midrule
\textit{Total natural} & --- & \textit{5{,}399} & --- & \textit{100.0} \\
\bottomrule
\end{tabular}
\end{table}

\FloatBarrier
\subsection{Evaluation Prompts}
\label{sec:app_prompts}

Each prompt is the concatenation of a task-specific base question and a format-specific output suffix.

\paragraph{Base question.}
\begin{itemize}\itemsep0.05em
    \item \textbf{Relations:} \texttt{Where is the \{object1\} in relation to the \{object2\}?}
    \item \textbf{Localization:} \texttt{Where is the \{object1\} in the image?}
    \item \textbf{Recognition (Shapes):} \texttt{What is the shape of the object in the image?}
    \item \textbf{Recognition (COCO):} per-sample question taken from the dataset's \texttt{question\_text} field, originally collected by \citet{neo2024towards}. Most pairs use \texttt{What is the person holding?}; the remainder cover variants such as \texttt{What is on the table?} or \texttt{What is the child holding?}. We prefill the assistant response with \texttt{It is a}.
\end{itemize}

\paragraph{Output suffix.}
\begin{itemize}
    \item \textit{choices}: \texttt{\{base question\} Answer only with $a_1, \dots, a_k$.}
    \item \textit{open}: \texttt{\{base question\} Answer with a single word.}
\end{itemize}
The options set $\{a_i\}$ is the sorted list of valid answers for the dataset (Table~\ref{tab:app_answer_spaces}). Under \emph{choices}, accuracy is the top-1 predicted token over the full vocabulary, while for \emph{open}, the argmax is restricted to the valid-answer logits. The \emph{open} protocol is our primary evaluation.

\begin{table}[htbp]
\centering
\caption{Answer space per dataset. The set serves as the prompted candidate list under \emph{choices} and as the constrained-logit set under \emph{open}.}
\label{tab:app_answer_spaces}
\small
\begin{tabular}{lll}
\toprule
Dataset & Task & Answer set \\
\midrule
Shapes Recognition  & recognition  & \{circle, square, triangle, star\} \\
Shapes Localization & localization & \{bottom, left, right, top\} \\
Shapes Relations    & relations    & \{above, below, left, right\} \\
\midrule
Controlled CLEVR        & relations    & \{behind, front, left, right\} \\
Controlled Images       & relations    & \{left, on, right, under\} \\
COCO\_two               & relations    & \{above, below, left, right\} \\
VG\_QA\_two             & relations    & \{behind, front, left, right\} \\
COCO\_one               & localization & \{bottom, left, right, top\} \\
VG\_QA\_one             & localization & \{bottom, left, on, right, top\} \\
VSR                     & relations    & \{above, behind, below, front, left, right\} \\
\midrule
COCO recognition        & recognition  & \{backpack, banana, baseball bat, bird, book, bowl, carrot, cat, cell \\ & & phone, cup, dog, donut, frisbee, giraffe, hair drier, horse, hot dog, kite,\\ & &  laptop, remote, toothbrush, vase\}. \\
\bottomrule
\end{tabular}
\end{table}

\paragraph{Multi-token object labels.}
Labels for \textit{Recognition} can span multiple subword tokens (e.g.\ ``baseball bat'' tokenizes as two pieces). For the constrained-logit evaluation under the \emph{open} protocol and the per-pair recognition comparison \(p(\text{object}_a) > p(\text{object}_b)\), we use the first token of each candidate label as its representative.

\paragraph{VSR adaptation.}
The original VSR benchmark~\citep{liu2023visual} uses sentence-level true/false judgments. We restrict to \texttt{label=True} samples and map the relation phrase to one of six directions via synonym groups: \emph{above} (``above'', ``on top of'', ``on'', ``over''), \emph{below} (``below'', ``under'', ``beneath''), \emph{left} (``left of'', ``at the left side of''), \emph{right} (``right of'', ``at the right side of''), \emph{behind} (``behind'', ``at the back of''), \emph{front} (``in front of''). Samples outside this map are discarded; retained samples use the relations template with the six-direction answer set.

\paragraph{Token-group definitions.} The pathway sets $\mathcal{I}$ and $\mathcal{T}$ (\S\ref{sec:experimental_setup}) are defined over input positions. The image set $\mathcal{I}$ contains only the visual patch tokens lying between the model's image-delimiter tokens (\texttt{<|vision\_start|>}/\texttt{<|vision\_end|>} for Qwen3 VL, \texttt{<img>}/\texttt{</img>} for InternVL3, and the contiguous \texttt{<image>} span for LLaVA-1.5). Delimiter tokens are not included in $\mathcal{I}$. The text set $\mathcal{T}$ contains only the natural-language query positions and excludes all model-specific chat-template and special tokens (e.g., role markers such as \texttt{<|im\_start|>}/\texttt{<|im\_end|>}). The final token $x_n$ forms its own group and is excluded from $\mathcal{T}$. Boundary and special tokens are never patched in any pathway intervention, so each pathway isolates a semantically meaningful region.

\FloatBarrier
\subsection{Dataset Samples}
\begin{figure}[htbp] %
    \centering

    \begin{subfigure}{\textwidth}
        \centering
        \includegraphics[width=0.95\linewidth]{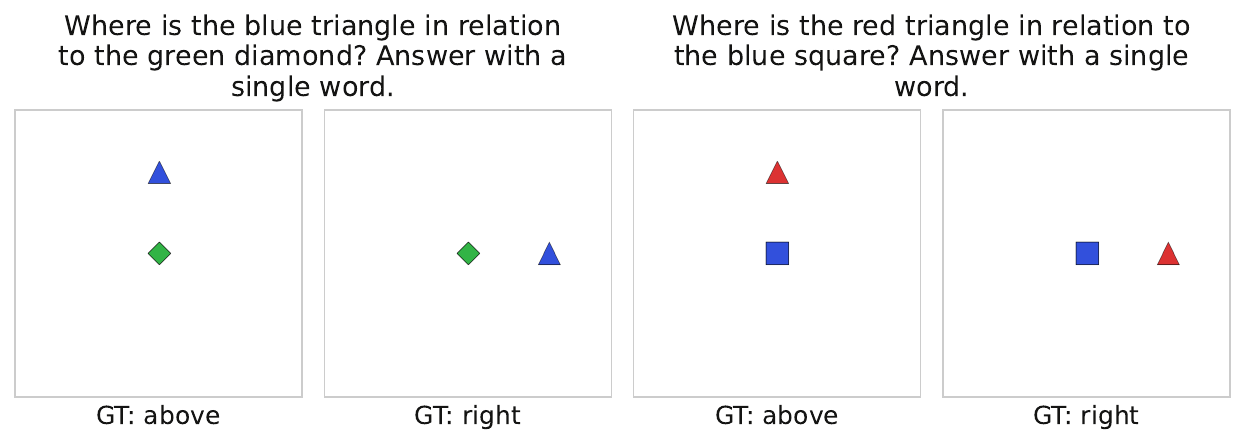}
        \caption{Samples of Shapes Relations.}
        \label{fig:samples_shapes_relations}
    \end{subfigure}
    
    \vspace{1.5em} %
    
    \begin{subfigure}{\textwidth}
        \centering
        \includegraphics[width=0.95\linewidth]{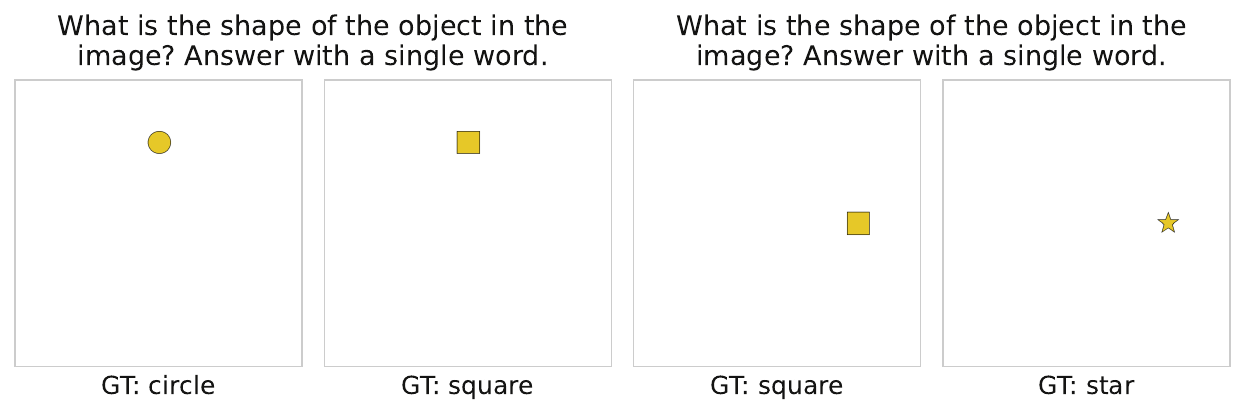}
        \caption{Samples of Shapes Recognition.}
        \label{fig:samples_shapes_recognition}
    \end{subfigure}
    
    \vspace{1.5em}
    
    \begin{subfigure}{\textwidth}
        \centering
        \includegraphics[width=0.95\linewidth]{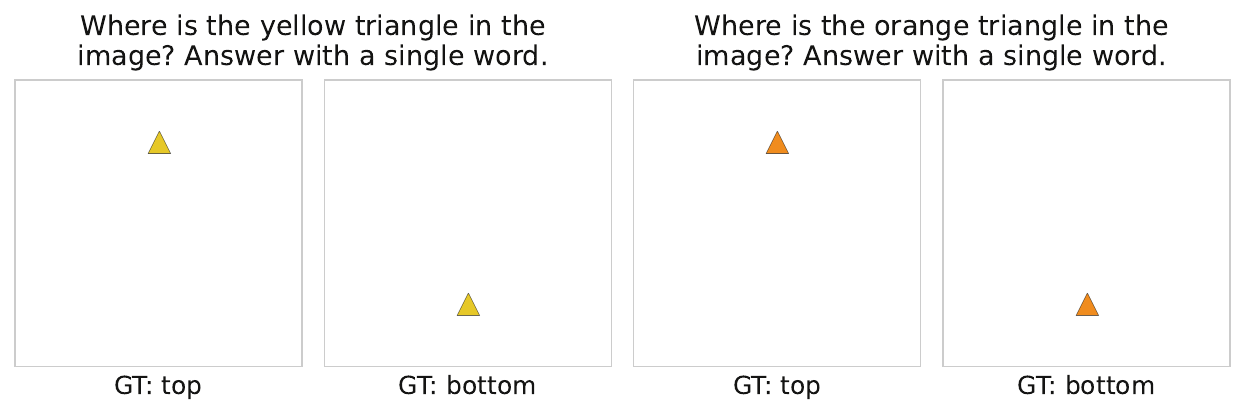}
        \caption{Samples of Shapes Localization.}
        \label{fig:samples_shapes_localization}
    \end{subfigure}
    
    \vspace{1.5em}
    \caption{Examples from the synthetic datasets. Each sample is paired with a counterfactual example used for causal patching, enabling controlled interventions on how visual information affects the final prediction across recognition, localization, and spatial relation tasks.}
    \label{fig:samples_syn_stacked}
\end{figure}

\begin{figure}[htbp] %
    \centering

    \begin{subfigure}{\textwidth}
        \centering
        \includegraphics[width=0.95\linewidth]{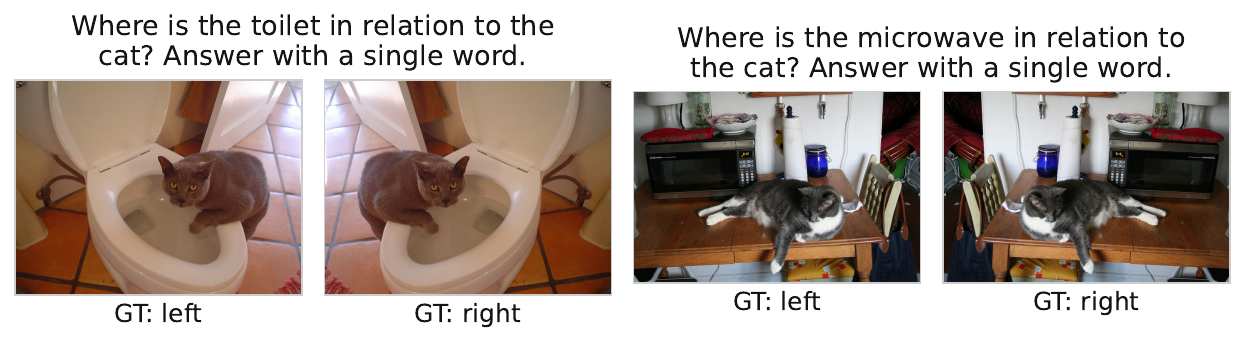}
        \caption{Samples of Natural Relations.}
        \label{fig:samples_natura_relations}
    \end{subfigure}
    
    \vspace{1.5em} %
    
    \begin{subfigure}{\textwidth}
        \centering
        \includegraphics[width=0.95\linewidth]{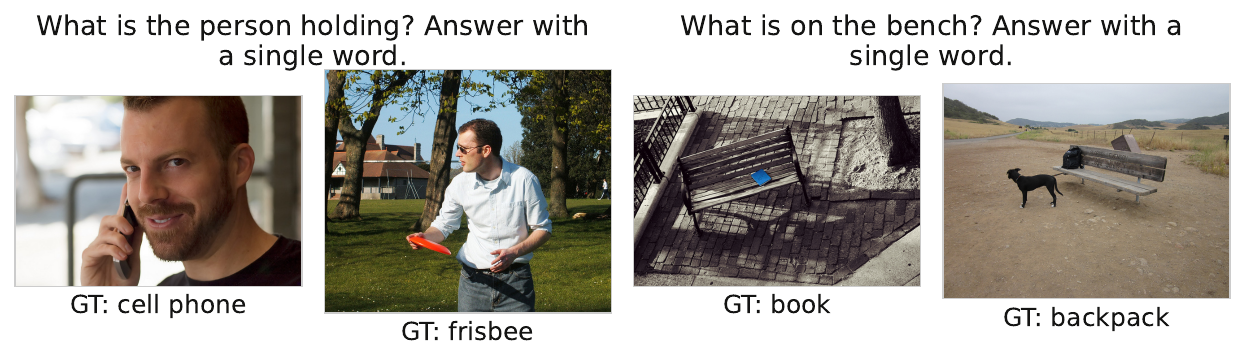}
        \caption{Samples of Natural Recognition.}
        \label{fig:samples_natural_recognition}
    \end{subfigure}
    
    \vspace{1.5em}
    
    \begin{subfigure}{\textwidth}
        \centering
        \includegraphics[width=0.95\linewidth]{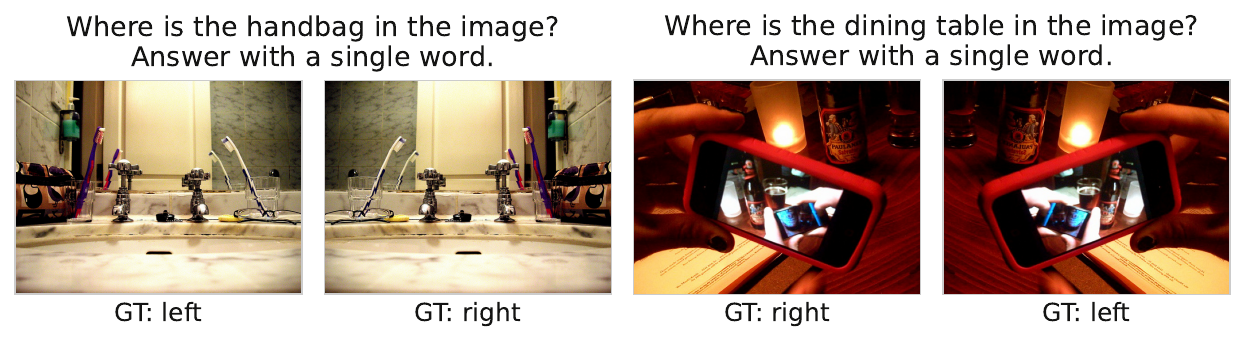}
        \caption{Samples of Natural Localization.}
        \label{fig:samples_natural_localization}
    \end{subfigure}
    
    \vspace{1.5em}
    \caption{Examples from the natural datasets. Each sample is paired with a counterfactual example used for causal patching, enabling controlled interventions on how visual information affects the final prediction across recognition, localization, and spatial relation tasks.}
    \label{fig:samples_nat_stacked}
\end{figure}

Figures \ref{fig:samples_syn_stacked} and \ref{fig:samples_nat_stacked} show samples of images from the synthetic and natural datasets, respectively, across the included tasks. Each sample includes a counterfactual, as used in causal patching.

\FloatBarrier
\section{Extended Results}
\FloatBarrier
\subsection{Per-Dataset Results}
\label{sec:app_per_dataset}
The main-paper aggregates all datasets. Here we break-out the results and present them by individual dataset. Baselines are shown in Table~\ref{tab:app_per_dataset_baselines}.

\begin{table}[htbp]
\centering
\caption{Per-dataset baseline accuracy for Qwen3-VL-4B across prompt formats. 
``Choices'' denotes multiple-choice prompting and ``Open'' denotes free-form generation. 
``No-image'' replaces the visual input with a black image while preserving the text prompt. 
Near-chance performance in the no-image condition indicates that the model relies primarily on visual information rather than prompt priors.}\label{tab:app_per_dataset_baselines}
\small
\begin{tabular}{llrrr}
\toprule
Dataset & Task & Choices (\%) & Open (\%) & No-image (\%) \\
\midrule
Shapes Relations            & relations    & 100.0 & 100.0 & 25.0 \\
Shapes Localization         & localization & 100.0 & 100.0 & 24.0 \\
Shapes Recognition          & recognition  & 100.0 & 100.0 & 24.5 \\
\midrule
Controlled CLEVR            & relations    & 99.8 & 94.6  & 25.0 \\
Controlled Images           & relations    & 97.8  & 99.5  & 25.0 \\
COCO\_two                   & relations    & 78.6  & 76.8  & 36.6 \\
VG\_QA\_two                 & relations    & 84.4  & 79.9  & 19.1 \\
VSR                         & relations    & 59.1  & 69.5  & 22.1 \\
\midrule
COCO\_one                   & localization & 68.4  & 55.6  & 25.2 \\
VG\_QA\_one                 & localization & 79.3  & 55.7  & 17.7 \\
\midrule
COCO recognition (paired)   & recognition  & 100.0 & 97.9  &  9.6 \\
\bottomrule
\end{tabular}
\end{table}

\begin{figure}[htbp]
  \centering
  \includegraphics[width=1\textwidth]{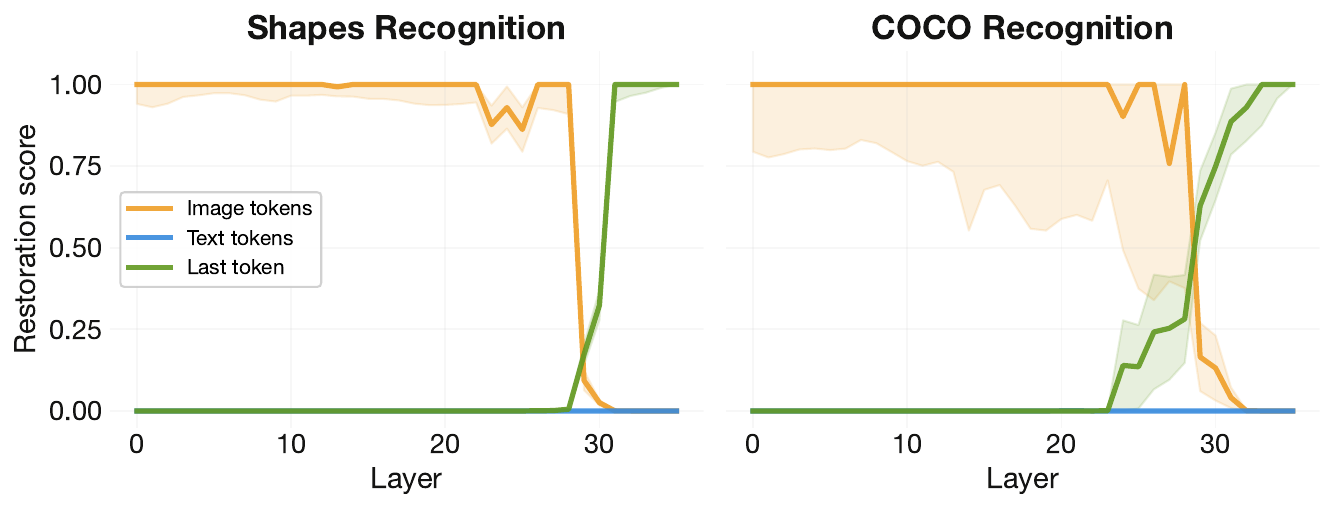}
  \caption{Per-dataset causal patching restoration on recognition (Qwen3-VL-4B). Both synthetic and natural recognition show the direct pathway with zero text restoration across all layers.}
  \label{fig:app_causal_recognition}
\end{figure}

\begin{figure}[htbp]
  \centering
  \includegraphics[width=1\textwidth]{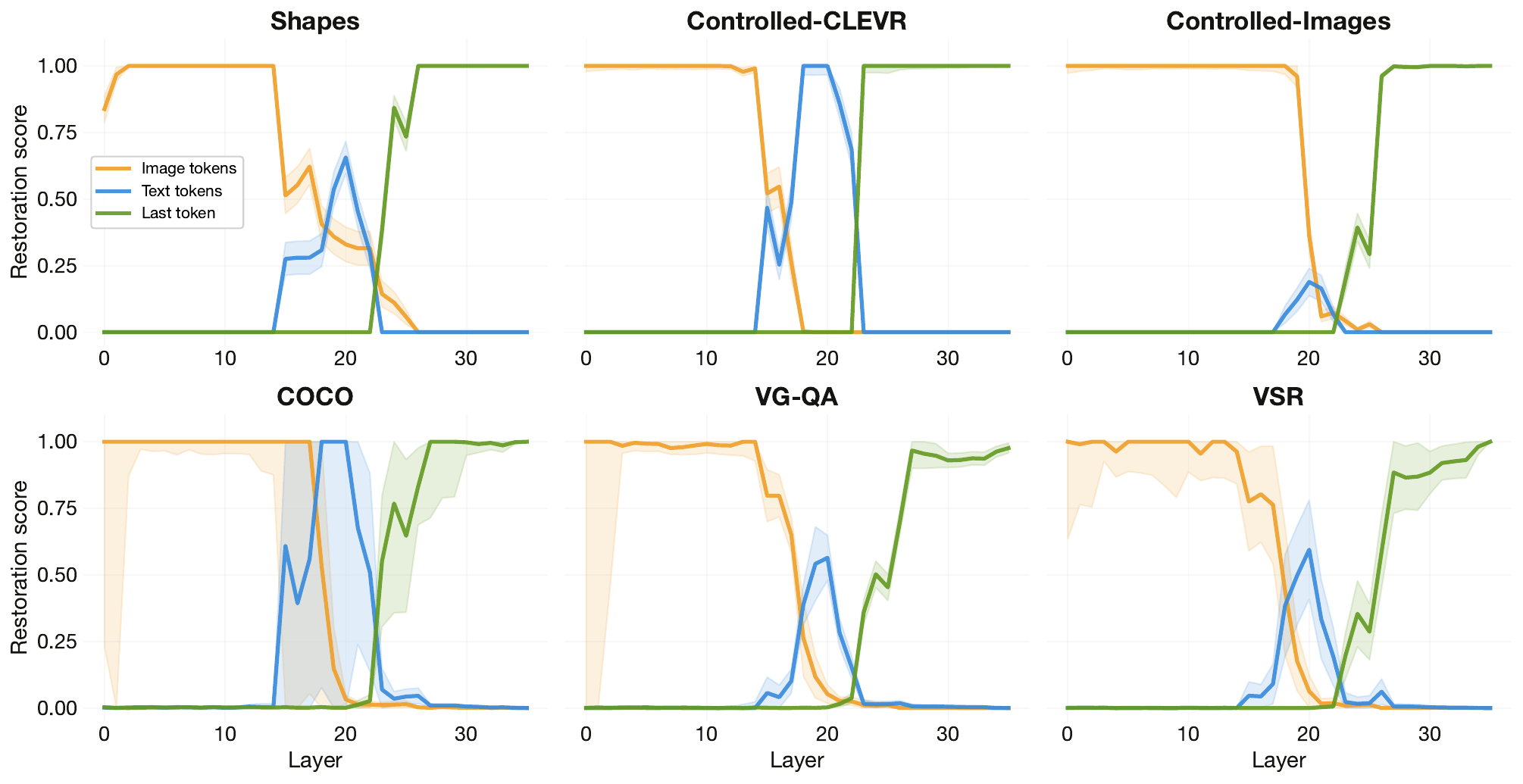}
  \caption{Per-dataset causal patching restoration on relations (Qwen3-VL-4B). Both synthetic and natural relations show the text mediated pathway.}
  \label{fig:app_causal_relations}
\end{figure}

\begin{figure}[htbp]
  \centering
  \includegraphics[width=1\textwidth]{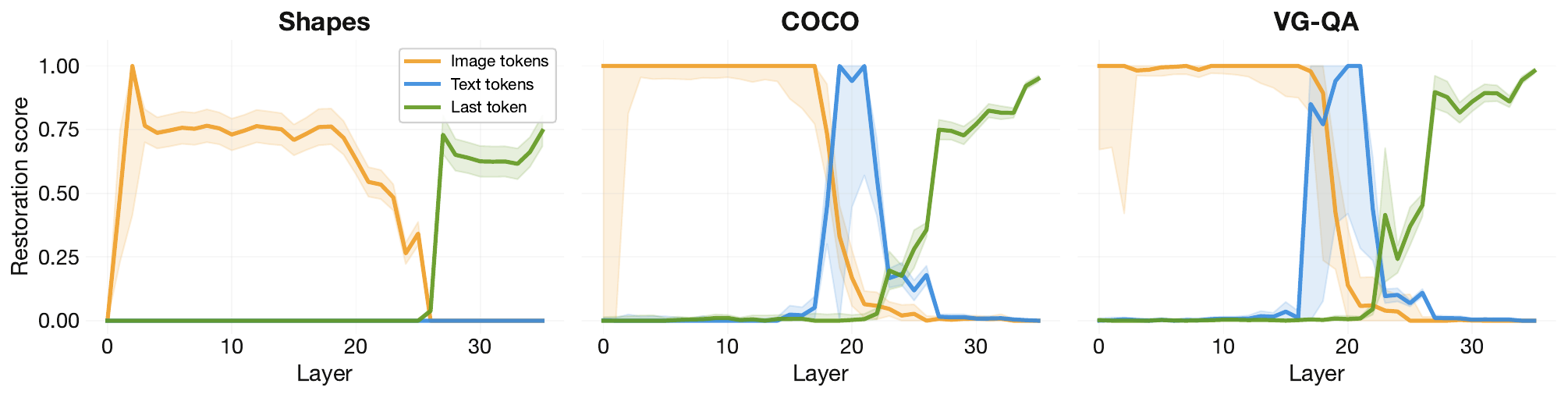}
  \caption{Per-dataset causal patching restoration on localization (Qwen3-VL-4B). Synthetic localization shows the direct pathway; the two natural datasets use text mediation.}
  \label{fig:app_causal_localization}
\end{figure}

\paragraph{Causal patching curves.}
Figures~\ref{fig:app_causal_recognition}, \ref{fig:app_causal_relations}, and~\ref{fig:app_causal_localization}, report per-dataset layer-wise restoration curves when patching image, query, or last-token hidden states on Qwen3-VL-4B. The aggregated patterns shown in the main paper hold consistently at the per-dataset level: relations show the three-stage image$\to$text$\to$last pipeline across all five datasets, recognition exhibits direct readout with negligible text restoration on both synthetic and natural data, and localization is data-dependent, direct on synthetic, mediated on the two natural datasets.

\paragraph{Corrupted-input text recovery.}
The main-paper recovery pattern holds at the per-dataset level (Table~\ref{tab:app_per_dataset_recovery}). Text-trajectory patching alone produces the correct answer at $\geq 81\%$ on natural data and 100\% on synthetic, including on recognition and synthetic localization where paired causal patching showed near-zero text restoration.

\begin{table}[htbp]
\centering
\caption{Per-dataset text recovery accuracy (Qwen3-VL-4B). Top-1 accuracy under all-layer text patching with the image replaced by uniform noise or black. $N_{\text{eval}}$: clean-correct samples used for the patched evaluation.}
\label{tab:app_per_dataset_recovery}
\small
\begin{tabular}{llrrr}
\toprule
\multirow{2}{*}{Dataset} & \multirow{2}{*}{Task} & \multirow{2}{*}{$N_{\text{eval}}$} & \multicolumn{2}{c}{Recovered (\%)} \\
\cmidrule(lr){4-5}
& & & noise & black \\
\midrule
Shapes Relations            & relations    & 400  & 100.0 & 100.0 \\
Shapes Localization         & localization & 400  & 100.0 & 100.0 \\
Shapes Recognition          & recognition  & 400  & 100.0 & 100.0 \\
\midrule
Controlled CLEVR            & relations    & 384  &  84.4 &  84.7 \\
Controlled Images           & relations    & 410  & 100.0 & 100.0 \\
COCO\_two                   & relations    & 338  &  98.2 &  99.1 \\
VG\_QA\_two                 & relations    & 229  &  98.7 &  97.8 \\
\midrule
COCO\_one                   & localization & 1{,}250 &  95.3 &  96.1 \\
VG\_QA\_one                 & localization &   647 &  96.8 &  98.0 \\
\midrule
COCO recognition (paired)   & recognition  &   144 &  81.9 &  91.7 \\
\bottomrule
\end{tabular}
\end{table}

\FloatBarrier
\Needspace{50\baselineskip}
\subsection{Top-1 predictions under patching}
\label{sec:app_argmax_outcomes}

\begin{figure}[H]
  \centering
  \includegraphics[width=1\textwidth]{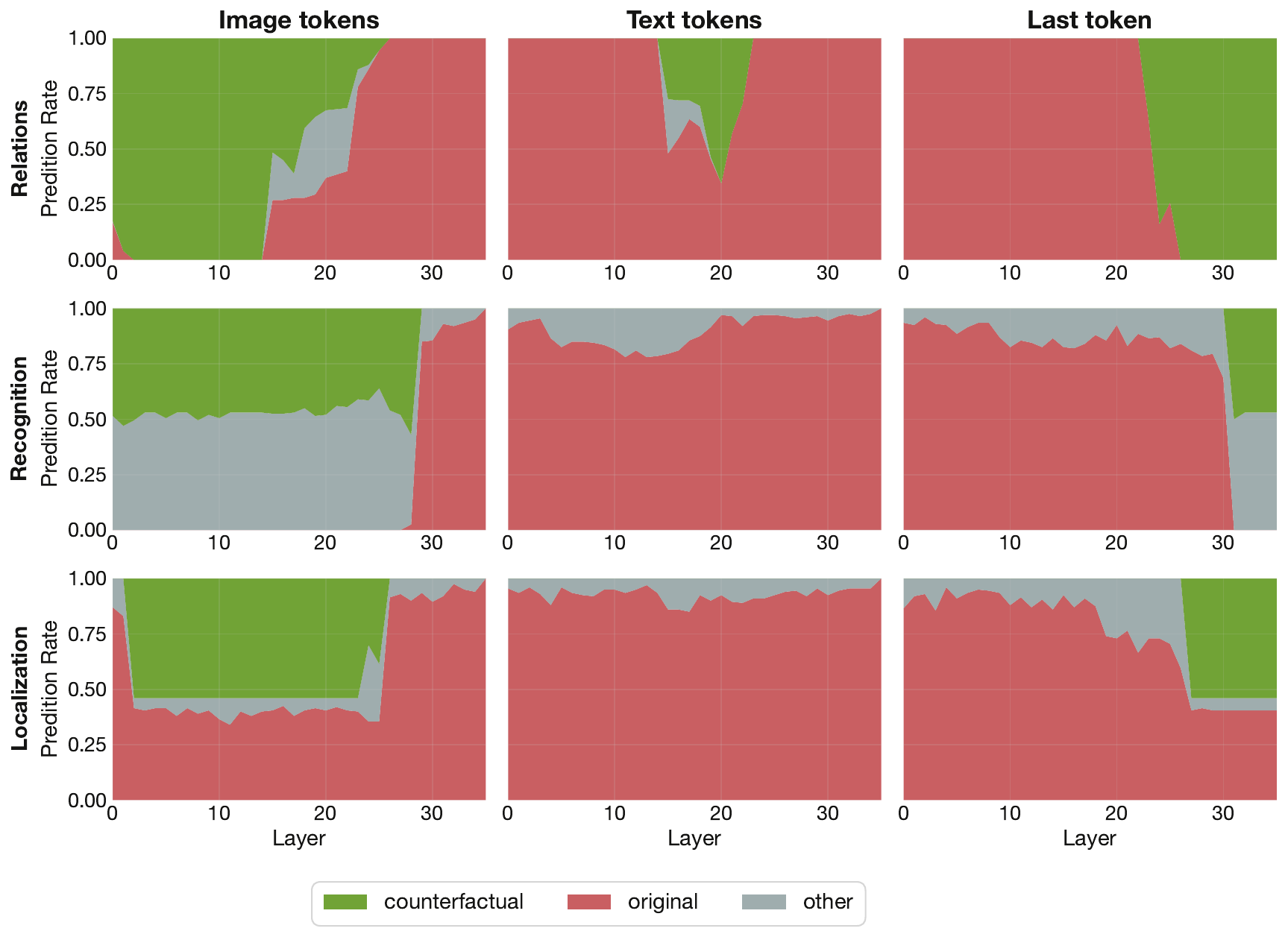}
  \caption{Argmax outcomes under patching for open prompts. For each layer and patched group, we report the fraction of samples whose top-1 prediction matches the original answer $y$, the counterfactual answer ${\tilde{y}}$, or another token. In the  open-generation setting, patched runs predominantly preserve either the original or counterfactual prediction, with a small fraction of out-of-vocabulary or paraphrased outputs.}
  \label{fig:argmax_outcomes_open}
\end{figure}

\begin{figure}[H]
  \centering
  \includegraphics[width=1\textwidth]{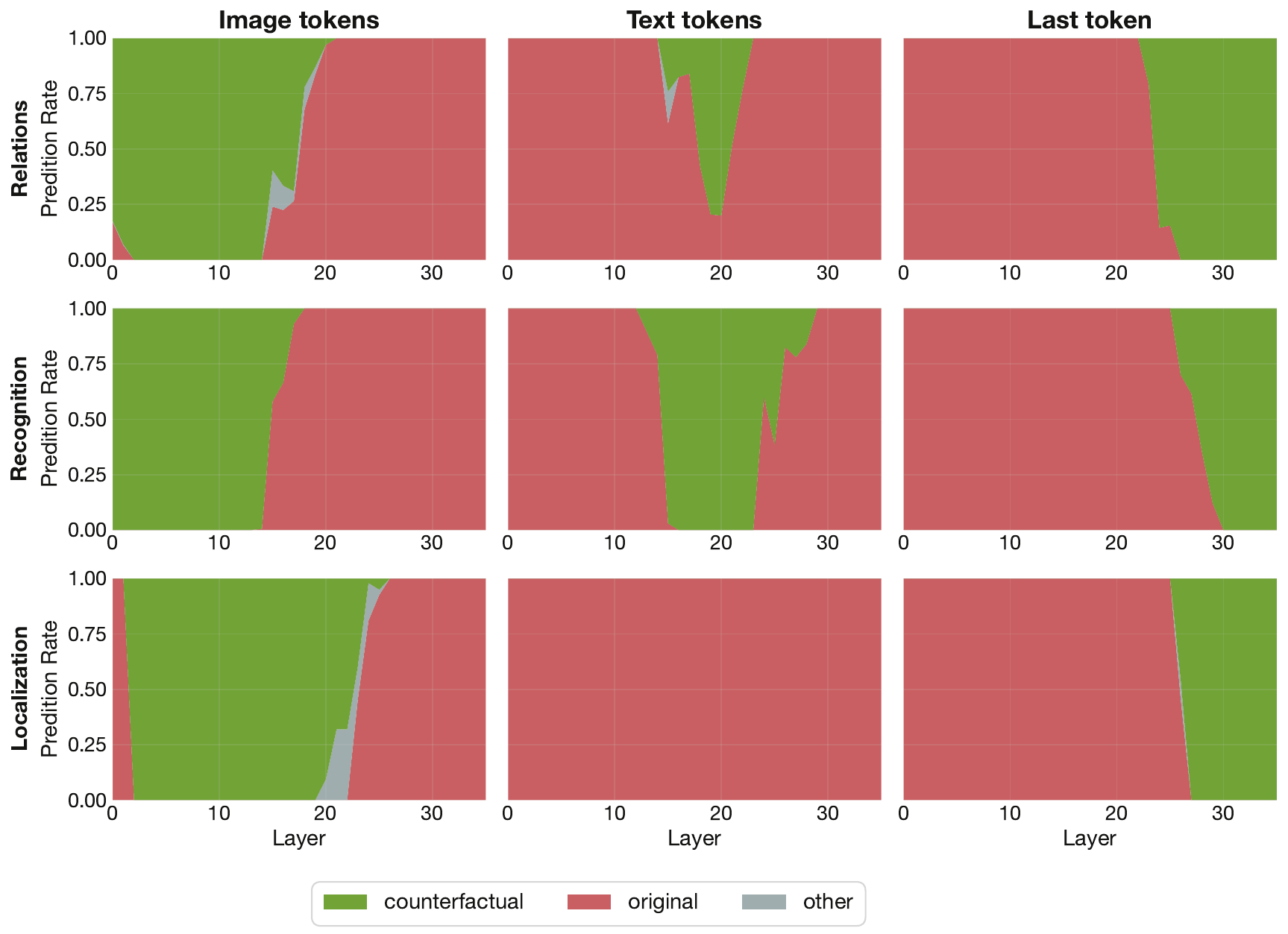}
  \caption{Argmax outcomes under patching for choices prompts. When candidate answers are explicitly listed in the prompt, patched runs almost always produce either the original answer $y$ or the counterfactual answer ${\tilde{y}}$, substantially reducing ``other'' outputs. Mid-layer text-token patching in recognition now produces a pronounced counterfactual band, consistent with prompt-dependent routing of object identity information into the text stream.}
  \label{fig:argmax_outcomes_choices}
\end{figure}

The restoration score introduced in \S\ref{sec:experimental_setup} (Eq.~\ref{eq:causal_score}) measures how much probability mass shifts toward the counterfactual answer under patching. We complement this metric with a discrete analysis: for every task, patched group, and layer, we store the fraction of evaluated pairs whose top-1 prediction matches (i) the counterfactual answer $\tilde{y}$, (ii) the original answer $y$, or (iii) any other token. Figures~\ref{fig:argmax_outcomes_open} and~\ref{fig:argmax_outcomes_choices} report these proportions for the \textit{open} and \textit{choices} prompts on the Shapes datasets (Qwen3-VL-4B).

Across tasks, patched groups, and layers, the ``other'' category is rarely dominant. Patching typically drives the argmax either toward $\tilde{y}$ (no transfer) or toward $\tilde{y}$ (successful transfer), rather than toward unrelated outputs. When candidate answers are explicitly listed in the prompt (Figure~\ref{fig:argmax_outcomes_choices}), the ``other'' category becomes even smaller: nearly every patched run produces one of the two expected answers. The open setting (Figure~\ref{fig:argmax_outcomes_open}) admits a small residual of out-of-vocabulary or paraphrased completions. 

These results also provide a discrete-prediction view of the prompt-modulation effect reported in \S\ref{sec:task_dependence} (``Prompt design can modulate pathway selection.''). For recognition, text-token patching under the \textit{open} prompt never flips the argmax to $\tilde{y}$, consistent with a direct readout in that setting. Under the \textit{choices} prompt, however, mid-layer text patching produces a substantial counterfactual band, indicating that the candidate-set prompt causes object identity information to become recoverable from the text pathway.

\FloatBarrier
\subsection{Answer Distributions Under Patching}
\label{sec:app_synthetic_loc_outputs}

Section~\ref{subsec:corrupted_patching} noted that synthetic shapes localization is an exception in the restoration-score metric: restoration is near zero, yet constrained-logit accuracy under patched text reaches 100\%. We resolve this apparent inconsistency here by inspecting unrestricted top-5 predictions on all 400 samples.

Table~\ref{tab:app_synth_loc_top1} reports the unrestricted top-1 token distribution in the three setups: (1) the input with the corrupted image, (2) the input with the clean image, and (3) the input with the corrupted image and the text patched in all layers. We observe that the patched run almost always outputs ``center'' as its top-1 token (399/400 samples), concentrating most of the probability mass on a response outside the candidate set $\{$top, bottom, left, right$\}$. This leaves little probability mass for the correct answer, resulting in a low restoration score. Under constrained-logit evaluation, however, the correct direction remains the highest-ranked candidate in 100\% of samples. The corrupted run (with no patching), in contrast, produces an unrelated high-frequency continuation token (``now'' in 100\% of samples), with the ground-truth direction never appearing in the unrestricted top-5.

The discrepancy between restoration score and constrained-logit accuracy therefore reflects the distinction between recovering the correct answer's probability and preserving its ranking among valid candidates. Although the probability assigned to the correct direction is not restored, its relative ranking indicates that the spatial information needed to distinguish between candidate directions remains available in the patched text representations.

\begin{table}[htbp]
\centering
\caption{Synthetic Shapes Localization, unrestricted top-1 distribution. Patched runs tend to generate ``center'' (a spatially coherent but out-of-vocabulary response). The inference from the corrupted input (with no patching) produces a non-spatial relevant token. With patching and restricted argmax, the model always answers correctly.}
\label{tab:app_synth_loc_top1}
\small
\begin{tabular}{lrrr}
\toprule
Top-1 token & Clean & Corrupted & Patched \\
\midrule
center      & 184 & 0   & 399 \\
left        & 108 & 0   & 0   \\
right       & 108 & 0   & 1   \\
now         & 0   & 400 & 0   \\
\midrule
GT in top-5 (unrestricted) & --- & 0 / 400 & 295 / 400 \\
GT = constrained argmax    & 400 / 400 & 0 / 400 & 400 / 400 \\
\bottomrule
\end{tabular}
\end{table}

\FloatBarrier
\subsection{Prompt-Format Modulation Across Tasks}
\label{sec:app_prompt_modulation}

\begin{figure}[htbp]
  \centering
  \includegraphics[width=1\textwidth]{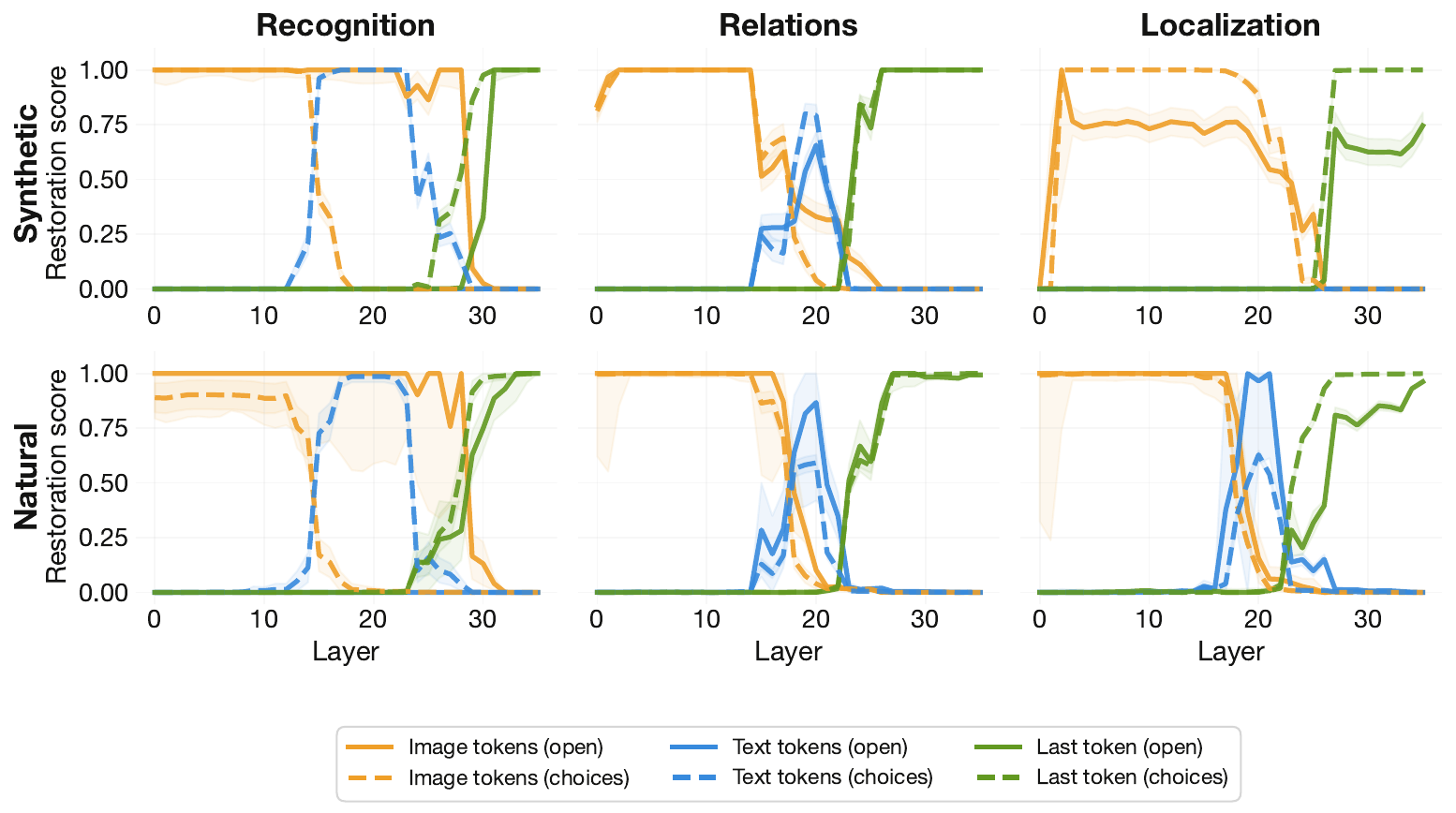}
  \caption{Prompt-format modulation across tasks (Qwen3-VL-4B). Layer-wise restoration under the \textit{open} (solid) and \textit{choices} (dashed) protocols, organized by task and data domain. Recognition is the only task where prompt format flips the dominant pathway. Relations and localization, prompt format modulates the magnitude of restoration but does not change which pathway dominates.}
  \label{fig:prompt_format_appendix}
\end{figure}

In Section~\ref{sec:task_dependence} we show that prompt format induces a within-task shift toward text-mediated routing for object recognition (Figure~\ref{fig:prompt_format}). We report the analogous comparison across all three task types here. Figure~\ref{fig:prompt_format_appendix} shows layer-wise restoration under the \textit{open} and \textit{choices} protocols for relations, localization, and recognition, on both synthetic and natural data.

Across relations and localization, the \textit{choices} protocol does not flip the dominant pathway: relations remain text-mediated under both protocols, and synthetic localization remains direct. The recognition flip described in the main paper is the only case where prompt format induces a qualitative pathway change with the same pattern holding for natural recognition data. 

\FloatBarrier
\subsection{Prompt-Paraphrasing Across Tasks}
\label{sec:app_prompt_paraphrasing}

To test whether pathway selection is sensitive to wording variation, we evaluate four additional prompt paraphrases per task. We perform causal patching of the text tokens at all layers, as in Table~\ref{tab:text_image_dominance}, and measure whether the model answers according to information in the image or text representations. Across all datasets and tasks, the dominant pathway remains unchanged (Tables~\ref{tab:pharaphrasing-relation}, \ref{tab:pharaphrasing-localization}, and~\ref{tab:pharaphrasing-recognition}). In contrast to task manipulation introduced in \S\ref{sec:task_dependence} where multiple-choice versus open generation alters the pathway selection for recognition, indicating that routing is sensitive to task formulation rather than wording variation.

\begin{table}[htbp]
\centering
\caption{Robustness to paraphrases on the Relations task. Percentage of predictions aligned with text or image information when all text tokens are patched in all layers.}
\label{tab:pharaphrasing-relation}
\resizebox{\columnwidth}{!}{%
\begin{tabular}{lllrr}
\toprule
\textbf{Prompt} & \textbf{Paraphrase Axis} & \textbf{Source} & \textbf{Image \%} & \textbf{Text \%} \\
\midrule
\multirow{2}{*}{Where is the \{A\} in relation to the \{B\}?} & \multirow{2}{*}{original} & Synthetic & 2.0 & 98.0 \\
 &  & Natural & 7.1 & 92.9 \\
\midrule
\multirow{2}{*}{Where is the \{A\} with respect to the \{B\}?} & \multirow{2}{*}{lexical} & Synthetic & 3.0 & 97.0 \\
 &  & Natural & 12.9 & 87.1 \\
\midrule
\multirow{2}{*}{State the position of the \{A\} relative to the \{B\}.} & \multirow{2}{*}{syntactic} & Synthetic & 13.5 & 86.5 \\
 &  & Natural & 6.3 & 93.7 \\
\midrule
\multirow{2}{*}{Looking at the image, where is the \{A\} compared to the \{B\}?} & \multirow{2}{*}{prefix + lexical} & Synthetic & 11.5 & 88.5 \\
 &  & Natural & 9.2 & 90.8 \\
\bottomrule
\end{tabular}%
}
\end{table}

\begin{table}[htbp]
\centering
\caption{Robustness to paraphrases on the Localization task. Percentage of predictions aligned with text or image information when all text tokens are patched in all layers.}
\label{tab:pharaphrasing-localization}
\resizebox{\columnwidth}{!}{%
\begin{tabular}{lllrr}
\toprule
\textbf{Prompt} & \textbf{Paraphrase Axis} & \textbf{Source} & \textbf{Image \%} & \textbf{Text \%} \\
\midrule
\multirow{2}{*}{Where is the \{A\} in the image?} & \multirow{2}{*}{original} & Synthetic & 100.0 & 0.0 \\
 &  & Natural & 10.2 & 89.8 \\
\midrule
\multirow{2}{*}{Where is the \{A\} located in the image?} & \multirow{2}{*}{lexical} & Synthetic & 100.0 & 0.0 \\
 &  & Natural & 6.5 & 93.5 \\
\midrule
\multirow{2}{*}{State the location of the \{A\} in the image.} & \multirow{2}{*}{syntactic} & Synthetic & 100.0 & 0.0 \\
 &  & Natural & 16.3 & 83.7 \\
\midrule
\multirow{2}{*}{Looking at the image, where is the \{A\}?} & \multirow{2}{*}{prefix + lexical} & Synthetic & 100.0 & 0.0 \\
 &  & Natural & 11.0 & 89.0 \\
\bottomrule
\end{tabular}%
}
\end{table}

\begin{table}[htbp]
\centering
\caption{Robustness to paraphrases on the Recognition task. Percentage of predictions aligned with text or image information when all text tokens are patched in all layers. For natural datasets we cannot use any paraphrase as the queries are sample specific.}
\label{tab:pharaphrasing-recognition}
\resizebox{\columnwidth}{!}{%
\begin{tabular}{lllrr}
\toprule
\textbf{Prompt} & \textbf{Paraphrase Axis} & \textbf{Source} & \textbf{Image \%} & \textbf{Text \%} \\
\midrule
What is the shape of the object in the image? & original & Synthetic & 100.0 & 0.0 \\
\midrule
Which object is shown in the image? & lexical & Synthetic & 100.0 & 0.0 \\
\midrule
Identify the shape of the object in the image. & syntactic & Synthetic & 100.0 & 0.0 \\
\midrule
Looking at the image, what is the shape of the object? & prefix + lexical & Synthetic & 100.0 & 0.0 \\
\midrule
Looking at the image, \{q\} & prefix & Natural & 95.1 & 4.9 \\
\bottomrule
\end{tabular}%
}
\end{table}

\FloatBarrier
\subsection{Prefilled Multi-Token Answers}
\label{sec:app_multitoken}
To test whether our findings extend beyond single-token answers, we perform an additional causal-patching experiment in which the model is prefilled with a partial answer, similar to the natural-recognition setup. We include the prefilled tokens in the text-token group $\mathcal{T}$ and patch $\mathcal{T}$ in all layers. The prompts and results are reported in Table~\ref{tab:prefilled_multitoken}. Across all tasks, we observe the same preferred pathway as in the original setup, suggesting that our conclusions are not specific to single-token answers. This controlled setup provides a proxy for multi-token generation, showing that the same two pathways are engaged at the critical prediction step when the model generates the answer-determining token.

\begin{table}[htbp]
\centering
\caption{All-layer text causal patching for \textbf{prefilled} prompts. Prefilled answer tokens are included in the patched text-token group as a proxy for multi-token generation.}
\label{tab:prefilled_multitoken}
\resizebox{\columnwidth}{!}{%
\begin{tabular}{llllcc}
\toprule
Task & Prompt & Prefill & Source & Image \% & Text \% \\
\midrule
Relations
& Where is the \{A\} in relation to the \{B\}?
& The \{A\} is located at the
& Synthetic & 0.0 & 100.0 \\
&&& Natural & 4.3 & 95.7 \\
\midrule
Localization
& Where is the \{A\} in the image?
& The \{A\} is located at the
& Synthetic & 100.0 & 0.0 \\
&&& Natural & 12.3 & 87.7 \\
\midrule
Recognition
& What is the shape of the object in the image?
& It is a
& Synthetic & 100.0 & 0.0 \\
& \{q\} (sample specific)
&& Natural & 96.7 & 3.3 \\
\bottomrule
\end{tabular}%
}
\end{table}

\FloatBarrier
\section{Robustness and Ablations}
\FloatBarrier
\subsection{Natural Counterfactuals}
\label{sec:app_hflips}
To enable causal patching on natural datasets, we construct counterfactual pairs using horizontal flips. We focus on samples from the relation and localization datasets where the answers depend on horizontal orientation (i.e., ``left'' or ``right''). For each such sample, we generate a horizontally flipped version of the image and update the ground-truth answer accordingly.

To ensure meaningful causal comparisons, we retain only pairs for which the model answers both the original and flipped samples correctly. This filtering step guarantees that both elements of the pair contain usable signal for causal patching, avoiding confounds due to model errors. Table~\ref{tab:hflip_baselines} reports baseline performance on the resulting horizontal-flip datasets. Horizontal flipping preserves geometry and task difficulty, the similar accuracy observed on original and flipped images confirms that the transformation introduces no difficulty.

\begin{table}[htbp]
\centering
\caption{Horizontal-flip baselines. Open-prompt accuracy (\%) of each model on the datasets used for causal patching with horizontal-flip counterfactuals. \emph{N Samples} is the total number of evaluation images. Each pair contributes both its original and horizontally flipped member, restricted to pairs whose original ground-truth answer is left'' or right''.}
\label{tab:hflip_baselines}
\resizebox{\textwidth}{!}{%
\begin{tabular}{llrrrr}
\toprule
Dataset & Task & N Samples & Qwen3-VL-4B (\%) & LLaVA-1.5-7B (\%) & InternVL3.5-4B (\%) \\
\midrule
Controlled CLEVR$_{\text{hflip}}$  & relations    & 408  & 100.0 & 100.0 & 100.0 \\
Controlled Images$_{\text{hflip}}$ & relations    & 412  & 99.0  & 96.4  & 99.0  \\
COCO\_two$_{\text{hflip}}$          & relations    & 558  & 97.7  & 81.7  & 95.0  \\
VG\_QA\_two$_{\text{hflip}}$        & relations    & 528  & 97.3  & 81.2  & 92.2  \\
\midrule
COCO\_one$_{\text{hflip}}$          & localization & 2280 & 95.2  & 87.0  & 89.6  \\
VG\_QA\_one$_{\text{hflip}}$        & localization & 1536 & 97.2  & 88.2  & 91.4  \\
VSR$_{\text{hflip}}$       & relations    & 74   & 90.5  & 73.0  & 81.1  \\
\bottomrule
\end{tabular}%
}
\end{table}

\FloatBarrier
\subsection{Corruption Ablation}
\label{sec:app_corruption_ablation}

In Section~\ref{subsec:corrupted_patching} we use uniform random pixel noise as the corrupted image input. We replicate the text-recovery experiment with two alternative uniform-color baselines (black and white) to verify the qualitative result is not sensitive to the choice of corruption.

\begin{table}[htbp]
\centering
\caption{Corruption-type ablation (Qwen3-VL-4B). Top-1 accuracy under all-layer text patching, aggregated per task across natural datasets, for three corruption types. Recovery is robust to the choice of corruption.}
\label{tab:app_corruption_ablation}
\small
\begin{tabular}{lrrr}
\toprule
Task                & Noise (\%) & Black (\%) & White (\%) \\
\midrule
Recognition         & 81.9 & 91.7 & 91.7 \\
Relations           & 94.9 & 95.1 & 94.7 \\
Localization        & 95.8 & 96.7 & 96.1 \\
\midrule
Synthetic (all)     & 100.0 & 100.0 & 100.0 \\
\bottomrule
\end{tabular}
\end{table}

Per-dataset breakdowns appear in Table~\ref{tab:app_per_dataset_recovery}. Across both noise and black corruptions, recovery rates are within $\sim$1 point on relations and localization, and the qualitative pattern is preserved: text-trajectory patching produces the correct answer in the large majority of cases on every task.

\FloatBarrier
\subsection{Windowed Causal Patching}
\label{sec:app_window_patching}
\begin{figure}[htbp]
\centering
\includegraphics[width=\textwidth]{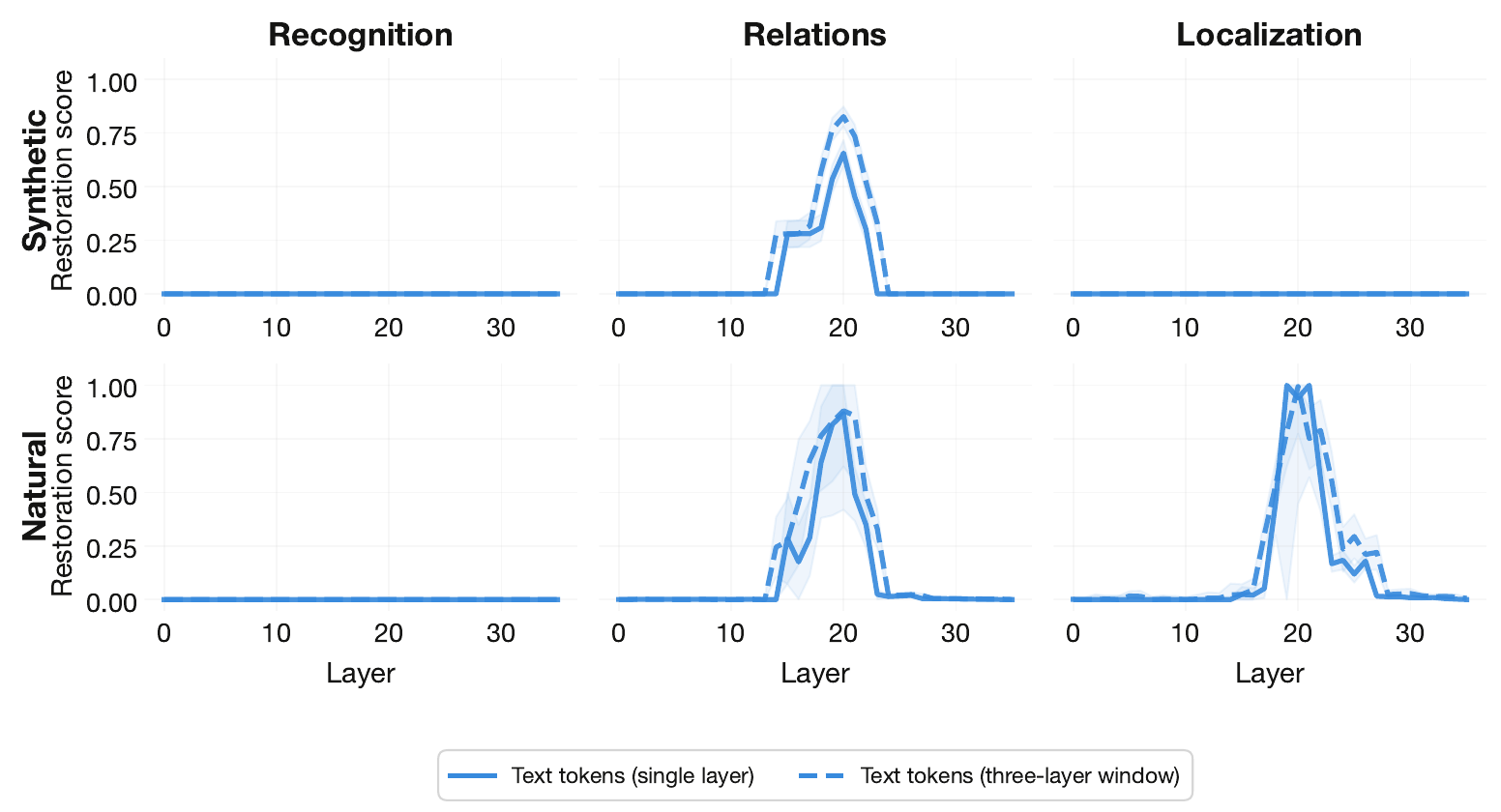}
\caption{Text-token causal patching with individual layers ($w=1$, solid) and contiguous three-layer windows ($w=3$, dashed) on Qwen3-VL-4B. Both settings preserve the prefered pathways.}
\label{fig:app_window_patching}
\end{figure}

Single-layer interventions might miss visual information that is distributed on neighboring layers. We repeat causal patching on Qwen3-VL-4B using a contiguous windows of three layers under the \textit{open} protocol. At each layer $\ell$, we patch the same token group at layers $\ell-1$, $\ell$, and $\ell+1$. Figure~\ref{fig:app_window_patching} compares query-text restoration under single-layer and three-layer interventions. The three-layer window preserves the pathways pattern reported in the main paper.

\FloatBarrier
\section{Generalization Across Models and Scales}
\label{sec:app_model_generalization}

In the main paper we present results on Qwen3-VL-4B. We replicate the core experiments on two additional architectures: LLaVA-1.5-7B and InternVL3.5-4B. We verify that our findings are not specific to a single model.

\FloatBarrier
\subsection{Causal Patching Across Model Families}
\begin{figure}[htbp]
  \centering
  \includegraphics[width=1\textwidth]{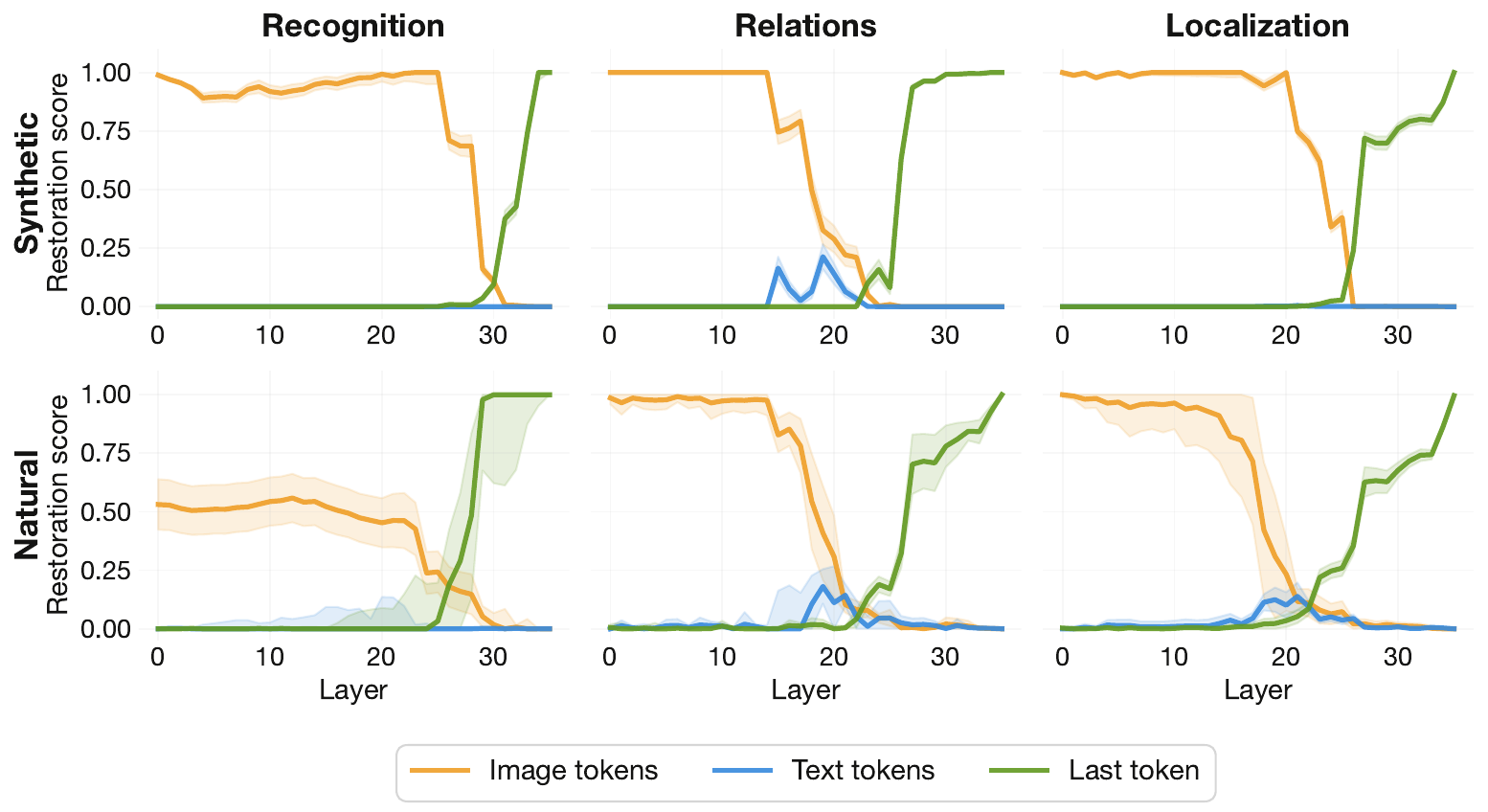}
  \caption{Causal patching restoration on InternVL3.5-4B.}
  \label{fig:causal_tracing_internvl}
\end{figure}

\begin{figure}[htbp]
  \centering
  \includegraphics[width=1\textwidth]{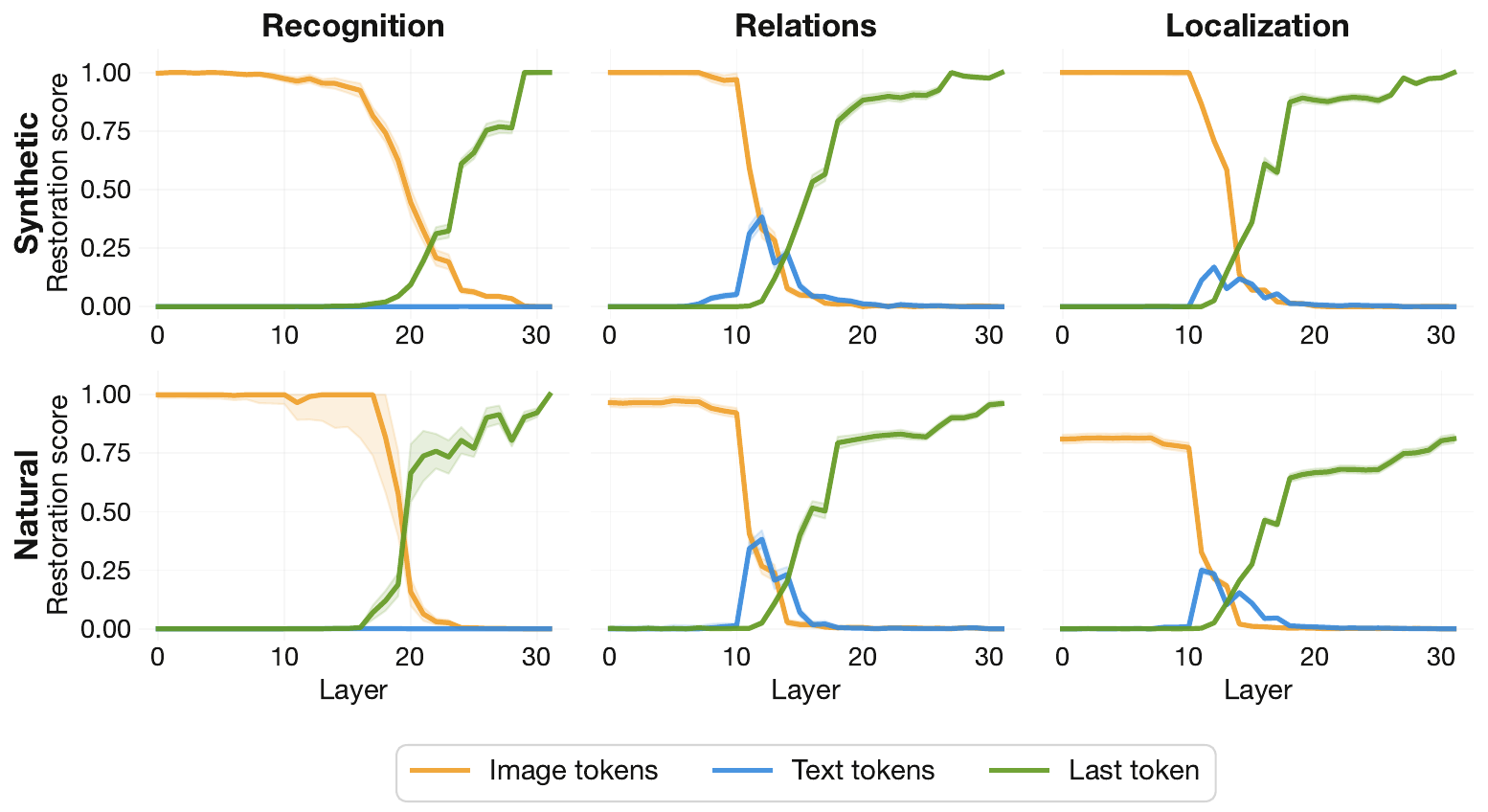}
  \caption{Causal patching restoration on LLaVA-1.5-7B.}
  \label{fig:causal_tracing_llava}
\end{figure}

Figures~\ref{fig:causal_tracing_internvl} and~\ref{fig:causal_tracing_llava} show layer-wise causal patching restoration for InternVL3.5-4B and LLaVA-1.5-7B, with panels organized by task in parallel to the Qwen results in the main paper.

\textbf{InternVL3.5-4B} reproduces the qualitative pattern observed on Qwen with somewhat lower text-token restoration: relations exhibit the three-stage image$\to$text$\to$last pipeline; recognition shows no text mediation; localization is data-dependent, with text mediation absent on synthetic data and present on natural data.

\textbf{LLaVA-1.5-7B} shows the same overall pattern, with one notable difference: text mediation is present on localization in both synthetic and natural data. Both InternVL and LLaVA produce somewhat lower restoration scores than Qwen across the board, but all three models exhibit a similar qualitative routing structure, with two coexisting pathways whose dominance depends on task and data.

\FloatBarrier
\subsection{Attention knockouts across model families}

Tables~\ref{tab:knockout_results}, \ref{tab:knockout_results_llava}, and~\ref{tab:knockout_results_internvl} report per-dataset knockout deltas for each model. Across all three architectures, blocking the direct pathway leaves performance largely intact, while blocking the mediated pathway causes substantial drops on tasks that rely on text mediation.

\begin{table}[htbp]
\centering
\caption{Attention knockout results on Qwen3-VL-4B. Each cell shows the difference in accuracy ($\Delta\%$) when blocking a specific attention path compared to the baseline.}
\label{tab:knockout_results}
\resizebox{\textwidth}{!}{
\begin{tabular}{llrrrrr}
\toprule
Dataset & Task & Base (\%) & NoImg (\%) &
$r_{\mathcal{I}\nrightarrow\text{last}}$ ($\Delta\%$) &
$r_{\mathcal{I}\nrightarrow\mathcal{T}}$ ($\Delta\%$) &
$r_{\mathcal{T}\nrightarrow\text{last}}$ ($\Delta\%$) \\
\midrule
Shapes Relations    & relations    & 100.0 & 25.0 & 0.0  & -10.5 & 0.0  \\
Shapes Localization & localization & 100.0 & 24.0 & 0.0  & 0.0   & 0.0  \\
Shapes Recognition  & recognition  & 100.0 & 24.5 & 0.0  & 0.0   & 0.0  \\
\midrule
Controlled CLEVR    & relations    & 94.6  & 25.0 & -9.6 & -41.2 & -6.9 \\
Controlled Images   & relations    & 99.5  & 25.0 & 0.0  & -40.5 & 0.0  \\
COCO\_two           & relations    & 76.8  & 36.6 & +1.6 & -27.5 & +0.7 \\
VG\_QA\_two         & relations    & 79.9  & 19.1 & +0.7 & -27.8 & +0.3 \\
COCO\_one           & localization & 55.6  & 25.2 & +2.0 & -17.0 & -0.7 \\
VG\_QA\_one         & localization & 55.7  & 17.7 & +6.8 & -20.9 & -3.5 \\
\midrule
VSR                 & relations    & 69.5  & 22.1 & +0.7 & -36.6 & -1.3 \\
\bottomrule
\end{tabular}
}
\end{table}

\begin{table}[htbp]
\centering
\caption{Attention knockout results on LLaVA-1.5-7B. Columns and convention as in Table~\ref{tab:knockout_results}.}
\label{tab:knockout_results_llava}
\resizebox{\textwidth}{!}{
\begin{tabular}{llrrrrr}
\toprule
Dataset & Task & Base (\%) & NoImg (\%) & $r_{\mathcal{I}\nrightarrow\text{last}}$ ($\Delta\%$) &
$r_{\mathcal{I}\nrightarrow\mathcal{T}}$ ($\Delta\%$) &
$r_{\mathcal{T}\nrightarrow\text{last}}$ ($\Delta\%$) \\
\midrule
Shapes Relations    & relations    & 28.5  & 24.8 & -0.2 & -3.7  & -0.2  \\
Shapes Localization & localization & 75.8  & 27.5 & -5.5 & -49.5 & -47.2 \\
Shapes Recognition  & recognition  & 100.0 & 24.5 & -0.5 & -75.5 & -16.5 \\
\midrule
Controlled CLEVR    & relations    & 67.4  & 25.0 & -2.7 & -42.4 & -38.2 \\
Controlled Images   & relations    & 29.1  & 25.0 & +1.0 & -4.1  & -4.1  \\
COCO\_two           & relations    & 55.2  & 30.7 & -0.9 & -26.1 & -22.7 \\
VG\_QA\_two         & relations    & 38.9  & 3.8  & +0.7 & -34.0 & -13.9 \\
COCO\_one           & localization & 46.4  & 25.6 & +0.3 & -21.1 & -19.6 \\
VG\_QA\_one         & localization & 16.3  & 0.2  & +4.4 & -15.5 & -9.4  \\
\midrule
VSR                 & relations    & 35.9  & 10.7 & -4.7 & -21.1 & -28.9 \\
\bottomrule
\end{tabular}
}
\end{table}

\begin{table}[htbp]
\centering
\caption{Attention knockout results on InternVL3.5-4B. Columns and convention as in Table~\ref{tab:knockout_results}.}
\label{tab:knockout_results_internvl}
\resizebox{\textwidth}{!}{
\begin{tabular}{llrrrrr}
\toprule
Dataset & Task & Base (\%) & NoImg (\%) & $r_{\mathcal{I}\nrightarrow\text{last}}$ ($\Delta\%$) &
$r_{\mathcal{I}\nrightarrow\mathcal{T}}$ ($\Delta\%$) &
$r_{\mathcal{T}\nrightarrow\text{last}}$ ($\Delta\%$) \\
\midrule
Shapes Relations    & relations    & 100.0 & 28.2 & 0.0   & -37.0 & 0.0  \\
Shapes Localization & localization & 100.0 & 23.0 & -1.5  & 0.0   & -9.0 \\
Shapes Recognition  & recognition  & 100.0 & 26.5 & 0.0   & 0.0   & 0.0  \\
\midrule
Controlled CLEVR    & relations    & 81.6  & 25.0 & -2.9  & -18.6 & +1.2 \\
Controlled Images   & relations    & 98.8  & 25.0 & -0.5  & -27.2 & -0.5 \\
COCO\_two           & relations    & 73.9  & 36.1 & -0.7  & -23.4 & -3.4 \\
VG\_QA\_two         & relations    & 65.3  & 19.8 & -2.8  & -25.0 & -0.3 \\
COCO\_one           & localization & 51.9  & 26.0 & +5.5  & -15.0 & -0.5 \\
VG\_QA\_one         & localization & 27.6  & 19.5 & +13.6 & -18.9 & +3.4 \\
\midrule
VSR                 & relations    & 40.9  & 8.7  & -2.7  & -18.8 & -1.3 \\
\bottomrule
\end{tabular}
}
\end{table}

\FloatBarrier
\subsection{Scale Generalization}
\label{sec:app_scale_generalization}
We extend all-layer text-token causal patching to Qwen3-VL-8B and Qwen3-VL-32B. Table~\ref{tab:app_scale_generalization} reports the fraction of predictions following the image or text information for each task. Each model is evaluated on its own clean-correct pairs, so sample counts differ across scales.

The information source used to generate the answers, i.e., the information present in the image tokens or the text tokens, is consistent across the tested 4B–32B model sizes and follows the task-dependent pattern presented in the main paper.

\begin{table}[htbp]
\centering
\caption{All-layer text-token causal patching across Qwen3-VL scales. $N$ denotes evaluated counterfactual pairs. Image and Text report the percentage of predictions aligned with the corresponding source of information.}
\label{tab:app_scale_generalization}
\small
\begin{tabular}{lllrrr}
\toprule
Task & Source & Model & $N$ & Image (\%) & Text (\%) \\
\midrule
Recognition & Synthetic & 4B & 200 & 100.0 & 0.0 \\
 &  & 8B & 200 & 100.0 & 0.0 \\
 &  & 32B & 200 & 100.0 & 0.0 \\
\midrule
Recognition & Natural & 4B & 61 & 96.7 & 3.3 \\
 &  & 8B & 73 & 100.0 & 0.0 \\
 &  & 32B & 56 & 100.0 & 0.0 \\
\midrule
Relations & Synthetic & 4B & 200 & 2.0 & 98.0 \\
 &  & 8B & 200 & 20.5 & 79.5 \\
 &  & 32B & 200 & 0.0 & 100.0 \\
\midrule
Relations & Natural & 4B & 959 & 8.6 & 91.4 \\
 &  & 8B & 964 & 1.3 & 98.7 \\
 &  & 32B & 958 & 0.9 & 99.1 \\
\midrule
Localization & Synthetic & 4B & 200 & 100.0 & 0.0 \\
 &  & 8B & 200 & 100.0 & 0.0 \\
 &  & 32B & 181 & 100.0 & 0.0 \\
\midrule
Localization & Natural & 4B & 1770 & 10.1 & 89.9 \\
 &  & 8B & 1753 & 14.9 & 85.1 \\
 &  & 32B & 1725 & 5.9 & 94.1 \\
\bottomrule
\end{tabular}
\end{table}

\FloatBarrier
\section{Implementation Details}
\label{sec:app_compute}
All experiments are inference-only on open-weight VLMs. The main-paper experiments run on a single NVIDIA A100-SXM 64GB GPU (CUDA 12.2). The larger-model study additionally uses multi-GPU inference: the Qwen3-VL-32B recognition-prefill runs allocate four GPUs. Each experiment (per dataset, per model, per intervention) completes in a few hours, and total compute across the three model families and all datasets is on the order of a few GPU-days.

\end{document}